\documentclass[11pt]{article}

\usepackage{amsmath , geometry , graphicx , amssymb, amsthm}
\usepackage{bm}
\usepackage{subcaption}
\usepackage{float}
\usepackage{multirow}

\usepackage[style=authoryear,citestyle=authoryear,natbib=true,backend=bibtex]{biblatex}
\title{ Classifying States of the Hopfield Network with Improved Accuracy, Generalization, and Interpretability }
\author{ Hayden McAlister$^{1}$, Anthony Robins$^{1}$, Lech Szymanski$^{1}$ }
\date{
  $^{1}$School of Computing, University of Otago, Dunedin, New Zealand.
}

\begin{document}
\maketitle	
\pagebreak

\begin{abstract}
    We extend the existing work on Hopfield network state classification, employing more complex models that remain interpretable, such as densely-connected feed-forward deep neural networks and support vector machines. The states of the Hopfield network can be grouped into several classes, including learned (those presented during training), spurious (stable states that were not learned), and prototype (stable states that were not learned but are representative for a subset of learned states). It is often useful to determine to what class a given state belongs to; for example to ignore spurious states when retrieving from the network. Previous research has approached the state classification task with simple linear methods, most notably the stability ratio. We deepen the research on classifying states from prototype-regime Hopfield networks, investigating how varying the factors strengthening prototypes influences the state classification task. We study the generalizability of different classification models when trained on states derived from different prototype tasks --- for example, can a network trained on a Hopfield network with 10 prototypes classify states from a network with 20 prototypes? We find that simple models often outperform the stability ratio while remaining interpretable. These models require surprisingly little training data and generalize exceptionally well to states generated by a range of Hopfield networks, even those that were trained on exceedingly different datasets.
\end{abstract}

\section{Introduction}

The Hopfield network \citep{Hopfield1982} is an associative memory consisting of a weight matrix that encodes the memories of the network. The Hopfield network has been formalized over both continuous and discrete domains, although we will focus only on the discrete domain in this paper. Thankfully, it has been shown that the associative memory properties of the continuous network are equivalent to those of the discrete network\citep{Hopfield1984}. States in the Hopfield network are represented by bipolar vectors \(\xi \in \{-1,1\}^N\) for a network dimension \(N\). The weight matrix, \(\bm{W} \in \mathbb{R}^{N \times N}\), is learned from a set of states, \(\bm{\xi}\), in a simple manner:
\begin{equation}
    \label{Eqn: Hebbian Learning}
    \bm{W} = \sum_{\xi \in \bm{\xi}} \xi \otimes \xi,
\end{equation}
where \(\otimes\) is the outer product operator. This is the Hebbian learning rule \citep{Hebb1949} in which neurons that have simultaneously similar activities have their synaptic strengths increase and neurons that have simultaneously dissimilar activities have their synaptic strengths decrease (or, in the bipolar domain, have that strength become negative). Other learning rules exist, such as the Widrow-Hoff / Delta rule \citep{WidrowHoff1960}, Storkey rule \citep{Storkey1997}, and thermal perceptron learning rule \citep{Frean1992}, but we will focus only on the Hebbian where prototype formation has been analyzed rigorously \citep{McAlisterEtAl2024a}. 

The states used to learn the weight matrix are exactly the class of ``learned states'', and are typically attractors of the associative memory (the network will iterate an initial ``probe'' state towards an attractor). Of course, it has been well studied that the Hopfield network cannot always stabilize every learned state as an attractor. Some sets of learned states may result in cross-talk that prevent attractors from forming, e.g. from two states being too close to one another. More commonly, the number of learned states may exceed the capacity of the Hopfield network, which for Hebbian learning is around \(0.138N\). This result has been shown both theoretically and practically many times \citep{Hopfield1982,Hertz1991} including from the perspective of statistical physics \citep{AmitEtAl1985,AmitEtAl1985a, McElieceEtAl1987} in which the Hopfield network is equivalent to the long-range Sherrington-Kirkpatrick spin-glass model \citep{KirkpatrickSherrington1978}. When the capacity of the Hopfield network is exceeded the network undergoes a sudden phase transition and nearly all of the learned attractors are lost. Replacing these are the second class of states we are interested in: ``spurious states'', which are stable attractors of the associative memory but were not learned. Spurious states are often present when the number of learned states is \textit{below} the capacity, and almost always serve to degrade the performance of an associative memory. However, some spurious states are useful. ``Prototype states'' are technically spurious, being stable attractors that are not presented in the learned state set, but act as representatives of a large number of those learned states. By stabilizing prototype states the Hopfield network can represent numerous related learned states with a single attractor and still capture useful features about the set. This can be seen as a trade off between representation and capacity, letting the Hopfield network to represent far more states than the capacity would otherwise allow. 

Previous work \citep{McAlisterEtAl2024a} has explored prototype states in the Hopfield network under Hebbian learning and found prototype strength to be proportional to
\begin{equation}
    |\eta| (1 - 4p + 4p^2)
\end{equation} 
for a number of instances $|\eta|$ and Bernoulli noise parameter $p\in[0,0.5]$ --- more instances and less noise creates stronger prototypes. They also found a prototype capacity equal to the usual Hebbian capacity of $0.138N$, adding a dependence on the total number of prototypes learned. We may only say a network is in a regime (non-prototype or prototype) when in the far limits of the above factors; networks trained on a small number of learned states that are unrelated to one another are non-prototype regime, and those trained on a large number of states that share many features are prototype regime. In between, when only a moderate number of examples of a prototype are presented, or the shared features are weak, the network may exhibit either behavior --- or more likely, neither behavior, stabilizing only spurious attractors as the individual learned states are above the Hebbian capacity and prototype states are not reinforced strongly enough.  There is no definite cutoff between non-prototype regime and prototype-regime Hopfield networks.  We will investigate only the extremes such that we may discuss non-prototype and prototype regimes with rigor.

Distinguishing between the classes of states is a useful labor. For example, in the realm of sequential learning the pseudorehearsal technique \citep{Robins1995,RobinsMcCallum1998} retrieves states from the Hopfield network to rehearse during future tasks. If the spurious states could be filtered out this would reduce to rehearsal, and improve the efficacy of the process immensely. It is no surprise then that classification of these states has been explored previously. \citet{RobinsMcCallum2004} looked at learned and spurious states in relatively small Hopfield networks (\(N=32\)), developing an effective approach in the stability ratio. The energy of a state \(\xi\) in the Hopfield network is given by:
\begin{equation}
    \label{Eqn: Energy of State}
    E(\xi) = -\frac{1}{2} \xi \odot \bm{W} \xi,
\end{equation}
where \(\odot\) is the element-wise product, such that \(E(\xi) \in \mathbb{R}^N\), i.e. a vector with one energy for each neuron in the network. Neurons with negative energy are stable, and those with positive energy are unstable. If a state has any neuron with a positive energy, that state is unstable and will yet iterate towards a nearby attractor. The energy of a state is a vector, with indices associated with neurons of the state. By sorting the energy we destroy that association but obtain a more useful representation of state stability: the energy profile. The energy profile is a fingerprint of the state stability without information about the specific neuron ordering. Importantly, the energy profile is comparable between Hopfield networks --- it does not matter if the neuron indices are reordered, the sorted energy profile will still indicate if a state is stable.

Returning to \citet{RobinsMcCallum2004}, the stability ratio is given by taking the ratio of the summed energies of the \(k\) most stable neurons to the summed energies of the \(k\) most unstable neurons (the top and bottom \(k\) entries of the energy profile). \citeauthor{RobinsMcCallum2004} use \(k=0.1 N\). They find the stability ratio is significantly different between learned and spurious states allowing for a linear separation between the classes --- although the criterion for this separation shifted between trials and networks. See Figure \ref{Fig: Stability Ratio Task 05} for a replication of these results in non-prototype-regime Hopfield networks, as \citeauthor{RobinsMcCallum2004} studied. 

\citet{GormanEtAl2017} continued work on the stability ratio by showing that prototype states are also distinguishable from both learned and spurious when using the stability ratio as a linear separation method. However, the experiments of \citeauthor{GormanEtAl2017} used a significantly different prototype generation method that does not have the theoretical basis our work relies on. \citeauthor{GormanEtAl2017} employ a Gaussian window method for prototype generation, while we use the Bernoulli framework set out in \citet{McAlisterEtAl2024a}. Differences between our works and their implications are discussed in Section \ref{Section: Experiment Two}. 

\citet{Abe1993} proposed a mathematically rigorous method for distinguishing spurious states from learned states in the Hopfield network tasked with solving optimization problems \citep{HopfieldTank1985}, such as the traveling salesman problem, by enforcing a monotonic and differentiable output function for each neuron and analyzing the theoretical attractor size for stable states. This approach is technically fascinating and mathematically robust, although the optimization mode of operation for the Hopfield network but does not lend itself to prototype states. Unlike the autoassociative mode of operation which we have discussed so far, in which learned states are free to interact to create representatives in the energy surface, the optimization mode of operation defines the energy surface to ensure specific states are minima, hence no interactions or representative states are admitted.

In psychology and cognition, prototypes provide a theoretical basis for learning and classification in the human brain. Prototype theory, also known as category theory and prototype-category theory, was introduced by \citet{Rosch1973,RoschMervis1975}. Prototype theory postulates that humans learn a single representative for a set of related items, compared future items against all prototypes to quickly estimate the properties of the new item. \citet{Rosch1973,RoschMervis1975} performed experiments in which participants were given items to memorize --- one group were given items that were all unrelated to one another, and another group given items that shared some characteristics. They found that participants learned quicker when presented with tasks structured around categories compared to unstructured tasks. Similar work was presented earlier by \citet{PosnerKeele1968}, who described prototype learning as exposure to examples that are distortions of the prototype. In both lines of research, it was found that participants could quickly classify new examples based on the prototypes learned. Prototype theory has been used in research of other fields, such as linguistics \citep{Ross1972, Ross1973, Corbett1978, Sadock2006} and anthropology \citep{Randal1976, Kempton1978}. Other models of categorization exist \citep{Kruschke2008}, with the most direct opposition to prototype-theory being exemplar-theory \citep{HomaEtAl1981, Nosofsky2011}, in which classification is performed by comparison to some exemplar items rather than a learned prototype. Some proponents of exemplar-theory assert that categories are ill-defined in any useful context, which naturally restricts the study of prototypes to purely academic applications \citep{Neisser1967, HomaEtAl1981}. Work on prototype-theory continued in shoring up both theoretical and experimental results \citep{Homa1984, SmithMinda1998, SmithMinda1999, SmithMinda2000}, refuting the exemplar-theory claims. Models of prototype formation proved useful in this endeavor, the Hopfield network among them, with particular note being given to the biological plausibility of the model \citep{Hopfield1984,HopfieldTank1985,PersonnazEtAl1986,KrotovHopfield2021} and Hebbian learning rule \citep{Hebb1949,JourneEtAl2022}. 

An abstraction to the Hopfield network, the Dense Associative Memory \citep{KrotovHopfield2016}, allows for increased capacities and new learning dynamics by replacing the quadratic energy function with faster-growing polynomials. Further abstractions using exponential energy functions \citep{DemircigilEtAl2017} results in exponential capacities and provide links to the attention mechanism from transformer architectures \citep{VaswaniEtAl2017, RamsauerEtAl2021}. More relevant to our work, the Dense Associative Memory (or, the Modern Hopfield Network when equipped with an exponential energy and operating over a continuous domain) has been observed to undergo a feature-to-prototype transition where memory vectors naturally switch from encoding features of the learned data to encoding prototypes \citep{KrotovHopfield2016,KrotovHopfield2018}. The mechanism behind this switch is very different to the prototype formation we study in this work, and classifying states of the Dense Associative Memory is outside the scope of this paper. However, we can employ the Dense Associative Memory as a classifier of Hopfield network states, which we explore briefly in Section \ref{Section: DAM Classifier Interpretability}.

In this paper we expand on the previous literature by using non-linear models, including deep, densely connected, feed-forward neural networks and support vector machines \citep{CortesVapnik1995}, to classify the states of a Hopfield network. Beyond simply improving on the state of the art, we investigate the generalization of each classifier; can we use a classifier trained on states derived from one Hopfield network to classify states derived from another, even when those Hopfield networks are trained on very different datasets? If so, how many Hopfield network's worth of states are required to train a generalizable classifier? To answer these questions prematurely: yes, and fewer than 10. We find that even very shallow neural networks are capable of distinguishing the classes well, often better than the stability ratio. These results are shown in Section \ref{Section: Experiment One}. We also find that the support vector machine equipped with a linear kernel can also distinguish the classes of states well, although non-linear kernels can provide greater generalization at the cost of interpretability. Generalization between prototype-regime and non-prototype-regime Hopfield networks is investigated in Section \ref{Section: Experiment Five}. We explore how the classifier performance and generalizability is affected by the factors discussed in \citet{McAlisterEtAl2024a}: the number of prototypes, the Bernoulli noise coefficient in the prototype data, and the number of instances per prototype. Our experiments are performed over synthetic datasets, allowing for fine control over all of these factors. Finally, in Section \ref{Section: Interpretability} we interpret our models, demonstrating that the performance improvements are not at the cost of explainability.
\section{Experiments}

\subsection{Models and Datasets}

The dataset for the classifier models are energy profiles drawn from Hopfield networks that  themselves are trained on prototype-regime datasets, as discussed above. To give a more concrete description; after selecting an arbitrary set of prototype states we generate a number of instances of each prototype by applying a Bernoulli random vector to the prototype states. The Bernoulli parameter (constant across the vector) controls how similar the instances are to the prototype, with smaller parameters corresponding to states that are more similar to the prototype. These examples form the class of learned states. The Hopfield network is then trained with Hebbian learning (Equation \ref{Eqn: Hebbian Learning}), and is probed with \(10,000\) uniformly random probes. To probe the network, we initialize the network with a state and repeatedly applying the relaxation rule
\begin{equation}
    \xi_i \left(t+1\right) := \text{sign}\left(\sum_j \bm{W}_{ji} \xi_j\left(t\right)\right),
\end{equation}
until the state has a negative energy for all neurons. The neuron index to update \(i\) is selected randomly. This process is guaranteed to terminate, as the update will always decrease the energy of the neuron, hence the sum of Equation \ref{Eqn: Energy of State} is a monotonically decreasing value and is a Lyapunov function \citep{HahnEtAl1963} for the system \citep{Hopfield1984}. This behavior has also been explored from the perspective of the Sherrington-Kirkpatrick spin-glass \citep{AmitEtAl1985}. Once the probe states are stable, we say they have iterated to some attractor of the associative memory. If that attractor is not a prototype state nor one of the learned states we add it to the class of spurious states.

For each experiment (outside of Section \ref{Section: Experiment One}) training data consists of energy profiles of states and class labels of learned, prototype, and spurious. We draw energy profiles from 10 Hopfield networks for training. Testing data is similarly the energy profiles of states and the class labels, but drawn from 100 further Hopfield networks. We are interested in classifying the states of new Hopfield networks, not the states from just a single network (for which not enough training data would exist to make a meaningful model).

Our ``standard'' conditions for the Hopfield network consists of 20 prototypes with 100 instances each, generated with Bernoulli parameter \(p=0.2\). This results in 2000 learned states total. The number of learned states is much larger than the capacity of the Hopfield network, but we are only concerned with whether we can distinguish the corresponding prototypes from spurious attractors not stabilizing individual instances of those prototypes. The number of spurious states is clearly not fixed but is found to be comparable to the number of learned states. Our Hopfield network has a dimension \(N=256\) through each experiment. Previous literature \citep{RobinsMcCallum2004,GormanEtAl2017} has explored stable learned states against spurious states, which we repeat in Section \ref{Section: Experiment Five}, although this work focuses on Hopfield networks with prototype states present. We have made no further efforts to ensure that prototype states are evenly spaced in state space, or that learned states are not duplicated, as the high dimension of the state space makes overlap exceedingly rare.

Contemporary approaches to a relatively simple classification task like this would develop an exceedingly deep feed-forward neural network to mine any and all features from the energy profile, up to and including over-fitting the training data. This approach is not overly useful to our application despite the allure of an easy solution --- a deep neural network does not have the interpretability of the existing stability ratio classifier and is hence less useful for studies on human cognition, where knowledge about \textit{why} two classes of attractors are different is just as important as finding \textit{that} they are different. For this reason we restrict ourselves to relatively simple models (although, we also investigate somewhat deep neural networks in Section \ref{Section: Experiment One} for completion).

For our neural network classifiers, we use only fully-connected layers and restrict our models to \(256\) inputs and \(3\) outputs to match the Hopfield network dimension and number of classes (learned, prototype, and spurious) respectively. Neural network classifiers are built with PyTorch \citep{AnselEtAl2024} using a cross-entropy loss, ReLU activation \citep{Agarap2019}, trained with the Adam optimizer \citep{KingmaBa2017} with learning rate \(10^{-3}\). All other parameters use their default values. For each experiment (outside of Section \ref{Section: Experiment One}) we use a neural network with layer sizes $(256, 3)$, i.e. a linear mapping. We found that adding an L2 weight decay term to the optimization improved weight interpretability. Our neural network loss for a tuple of training data and label \((X, y)\), the network parameters \(\bm{W}\), and network function \(f(\cdot)\) is given by:
\begin{equation*}
    \mathcal{L}\left(X,y,\bm{W}\right) = \mathcal{L}_{\text{Cross-Entropy}}\left(f(X),y\right) + \lambda \sum_{w\in\bm{W}} w^2.
\end{equation*}
Our weight decay parameter is set to \(\lambda=10.0\) throughout these experiments, although a very wide range of values (\(1 - 10000\)) were tested and performed well. This value is relatively high in the context of neural network regularization, but we found this value gave a more interpretable set of parameters without degrading accuracy.

Support vector machine classifiers are built using scikit-learn \citep{scikit-learn, sklearn-api} using both the linear and radial basis function kernels, as well as a regularization parameter of \(C = 0.001\). We use the squared-hinge loss to train the SVM. The regularization parameter \(C\), playing a similar role to weight decay \(\lambda\) in the neural network, is found to be more important in the support vector machine for ensuring learned coefficients remain smooth. Also note that by convention the role of \(C\) is inverted compared to weight decay; \textit{smaller} \(C\) results in \textit{greater} regularization. The loss function of the support vector machine is given by:
\begin{equation*}
    \mathcal{L}\left(X,y,\bm{W}\right) = C\mathcal{L}_{\text{Squared-Hinge}}\left(f(X),y\right) + \sum_{w\in\bm{W}} w^2.
\end{equation*}

Our stability ratio model calculates the stability ratio using \(k= 0.1N\) as suggested by \citet{RobinsMcCallum2004}. For our networks, this corresponds to \(k = 25\) neurons. We train a multinomial logistic regression using the stability ratio, adding an L2 regularization term is applied with strength \(1.0\). Again, regularization was added and hyperparameters tuned to improve interpretability, see Section \ref{Section: Interpretability}.

For both the neural network and support vector machine classifier the input is the sorted energy profile of the states, and the labels are one-hot encoded class indices. Biologically, sorting is not an impossible feat. More likely, however, would be some kind of partial sort, or competitive network structure that can filter for the strongest and weakest stabilities. Considering the nature of the original stability ratio, which specifically calls for the most and least stable neurons, such a structure could theoretically implement a similar algorithm. The models we employ in this paper, relying more heavily on the full sorted vector of stabilities, may be less biologically plausible, but not to the extent of deeper, black-box models.  For all of our classifiers we introduce class weightings such that each class has equal influence on the parameter updates. 

\subsection{Energy Profile}

\begin{figure}[H]
    \centering
    \begin{subfigure}[t]{0.45\textwidth}
        \centering
        \includegraphics[width=\textwidth]{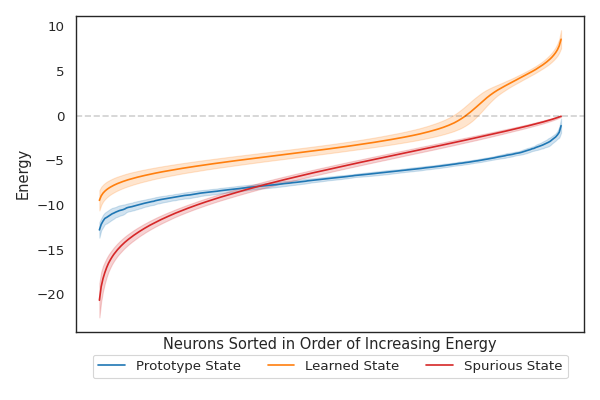}
        \caption{Non-normalized Energy Profiles.}
    \end{subfigure}
    \begin{subfigure}[t]{0.45\textwidth}
        \centering
        \includegraphics[width=\textwidth]{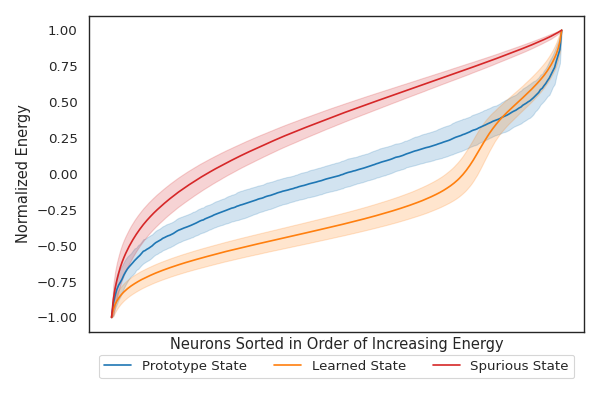}
        \caption{Normalized Energy Profiles.}
    \end{subfigure}
    \caption{Energy profiles of states from the standard conditions of the Hopfield network in both the normalized and non-normalized form. The mean energy profile is shown as a solid line, and the standard deviation of each neuron is shown as a shaded area around the mean. Note that the non-normalized energy profiles have an interpretable criterion at \(E=0\), where states with any energy above zero are unstable (all learned states) and those with all energy below zero are stable (prototypes and spurious).}
    \label{Fig: Example Energy Profiles}
\end{figure}

The energy profile is crucial to creating a generalizable classifier. Since the magnitude of the energy is determined by the magnitude of the weight matrix by Equation \ref{Eqn: Energy of State}, which is in turn influenced by the number of learned states, it would be sensible to remove this dependence by normalizing the energy profile, as not every Hopfield network is trained on the same number of learned states. Normalizing the energy profile involves scaling the elements of the sorted vector to have minimum value \(-1\) and maximum value \(1\). Figure \ref{Fig: Example Energy Profiles} shows the same energy profiles in both the non-normalized and normalized forms. We are not interested energy profiles that are normalized using statistics from the states of the respective Hopfield network i.e. scaling such that the maximum energy of all states from the same network is \(1\), as in practice we may not have access to every prototype, learned, and spurious state of our network to repeat this normalization. We have run our experiments with both the non-normalized energy profiles and normalized energy profiles. In general, we found better classifier performances with non-normalized energy profiles than normalized, and hence show results for classifiers trained on non-normalized energy profiles throughout this paper, although we will discuss any interesting findings from the normalized profiles when they arise. This is a somewhat surprising result, as we would expect normalized energy profiles to generalize better, but it seems that most classifiers take advantage of the context of the non-normalized energy profiles, e.g. that spurious states can never have a positive energy neuron by definition. Information like this that is removed by normalization is apparently more useful for classifiers (in most situations) and hardly seems to affect generalization at all.

The energy profiles in Figure \ref{Fig: Example Energy Profiles} are also the input data for our classifiers. Altering the dataset of the Hopfield network will ultimately alter the shape of the energy profiles, e.g. a lower Bernoulli parameter would lower the prototype energy as prototype states become more stable. It is for this reason we are focused on simple models, which remain interpretable when compared to more complex models. Although the current task is relatively straightforward and may be inspected directly we wish to develop models that can be applied to any Hopfield network no matter the makeup of the learned states. For example, the energy profiles and stability ratios found in \citet{GormanEtAl2017} are significantly different to those found in our work, likely due to their use of different prototype generation methods and learning rules. Our models need to be robust enough to handle any energy profile, even those that we have not yet observed, while still allowing for inspection of what features of those energy profiles are used in classification. 

Notice in Figure \ref{Fig: Example Energy Profiles} that the energy profiles have very small standard deviations. We suspect that neural network classifiers will learn hypotheses that only barely separate classes in the training data, moving decision boundaries incrementally and slowing considerably when those boundaries include all of the training instances. This would result in a classifier that is not very robust to significant changes to the energy profiles, such as a Hopfield network trained on a different dataset, and hence a classifier that does not generalize well. Cross-entropy loss helps avoid this problem somewhat, although we believe a maximum-margin model would sidestep this problem entirely, ensuring the decision boundaries leave the largest room for movement in the energy profiles. This motivates our inclusion of support vector machines in our model line-up, even though their interpretability could be better.

\subsection{Dataset Imbalance and Macro F1 Score}

We measure classifier performance using the macro F1 score, as the testing datasets have extreme imbalance. Since the number of learned states must necessarily be much larger than the number of prototype states, by a factor of \(100\) in our standard conditions, we must address the class imbalance in whatever measure we select. Accuracy alone would be insufficient, relegating the classification accuracy on prototype states to a small impact on the decimal places. We could report only the confusion matrix for each experiment, which accurately indicates classifier performance even on small classes, but cannot be effectively aggregated across many classifiers and is thus difficult to present in any statistically significant fashion.

\begin{table}[H]
    \centering
    \begin{tabular}{cc|ccc||c}
        \multicolumn{2}{c}{} & \multicolumn{3}{c}{Predicted Label} & \\
        \multicolumn{2}{c}{} & Prototype & Learned & Spurious & Accuracy\\
        \cline{3-6}
        \multirow{3}{*}{True Label} 
        & Prototype & $2869$ & $31$ & $100$ & $95.6\%$ \\
        & Learned   & $77582$ & $222304$ & $114$ & $74.1\%$ \\
        & Spurious  & $3683$ & $0$ & $434678$ & $99.2\%$ \\
    \end{tabular}
    \caption{A confusion matrix from a classifier applied to testing data drawn from the standard conditions. Note the extreme class imbalance.}
    \label{Table: Example Confusion Matrix}
\end{table}

\begin{table}[H]
    \centerline{
    \begin{tabular}{ c c }
    Metric & Value \\
    \hline
    Accuracy & 0.8901 \\
    Micro F1 Score & 0.8901 \\
    Macro F1 Score & 0.6375
    \end{tabular}
    }
    \caption{Aggregate metrics for the confusion matrix shown in Table \ref{Table: Example Confusion Matrix}.}
    \label{Table: Example Metrics}
\end{table}

In Table \ref{Table: Example Confusion Matrix} we see a confusion matrix typical of our experiments. Looking across the rows, the classifier is relatively accurate at predicting each class. The worst class, learned states, is classified about \(75\%\) accurately, meaning our model has high recall (in the statistical meaning) on all classes. However, looking down the columns it is clear that the classifier cannot be trusted to predict prototype states effectively. Fewer than \(5\%\) of the states predicted to be prototypes are truly prototypes, the remainder are misclassified from other classes, meaning our model has poor precision on prototype states. This does not significantly impact accuracy or micro F1 score since there are simply fewer prototype states than other classes, reflected in Table \ref{Table: Example Metrics}. Macro F1 score more accurately shows the poor performance on the test dataset. While it is true that all metrics discussed here would increase if the number of misclassified states were decreased, macro F1 score offers a broader range of values and is more sensitive for the data we are using, leading to more obvious changes in classifier performance than a relatively tiny increase in accuracy would. For this reason we present macro F1 score throughout this paper.
\section{Results}
\label{Section: Results}

Throughout these Sections we conduct experiments training classifiers on datasets that vary only a single variable from the standard conditions, allowing us to investigate the effect of that variable on classifier performance and generalization. In each experiment we test the classifier on each varied dataset as well: e.g. we test how a classifier trained on data with 20 prototypes performs on data with 10 prototypes and vice versa. We include an additional dataset in our training/testing, the ``combined'' dataset, in which all other datasets from that experiment are merged. When training on the combined dataset, the indicated number of Hopfield networks are drawn from \textit{each} other class, meaning the classifiers trained on the combined dataset will be exposed to more data overall than other classifiers. This is necessary as we use only a small number of Hopfield networks for training --- if we instead \textit{sample} from the other datasets, we would exclude some from our training data which would be misleading. In Section \ref{Section: Experiment One} we investigate the architecture of the classifiers, measuring the macro F1 score of each class for classifiers with increasing depth. Section \ref{Section: Experiment Two} we measure the macro F1 score of classifiers trained on Hopfield networks with varying numbers of prototypes. Section \ref{Section: Experiment Three} investigates the Bernoulli parameter, Section \ref{Section: Experiment Four} investigates the number of learned states, and Section \ref{Section: Experiment Five} looks at generalizability between prototype-regime and non-prototype-regime Hopfield networks. We show the stability ratios of each experiment in Appendix \ref{Appendix: Stability Ratios} 

\subsection{Neural Network Architecture}
\label{Section: Experiment One}

\begin{figure}[H]
    \centering
    \begin{subfigure}[t]{0.45\textwidth}
        \centering
        \includegraphics[width=1.0\textwidth]{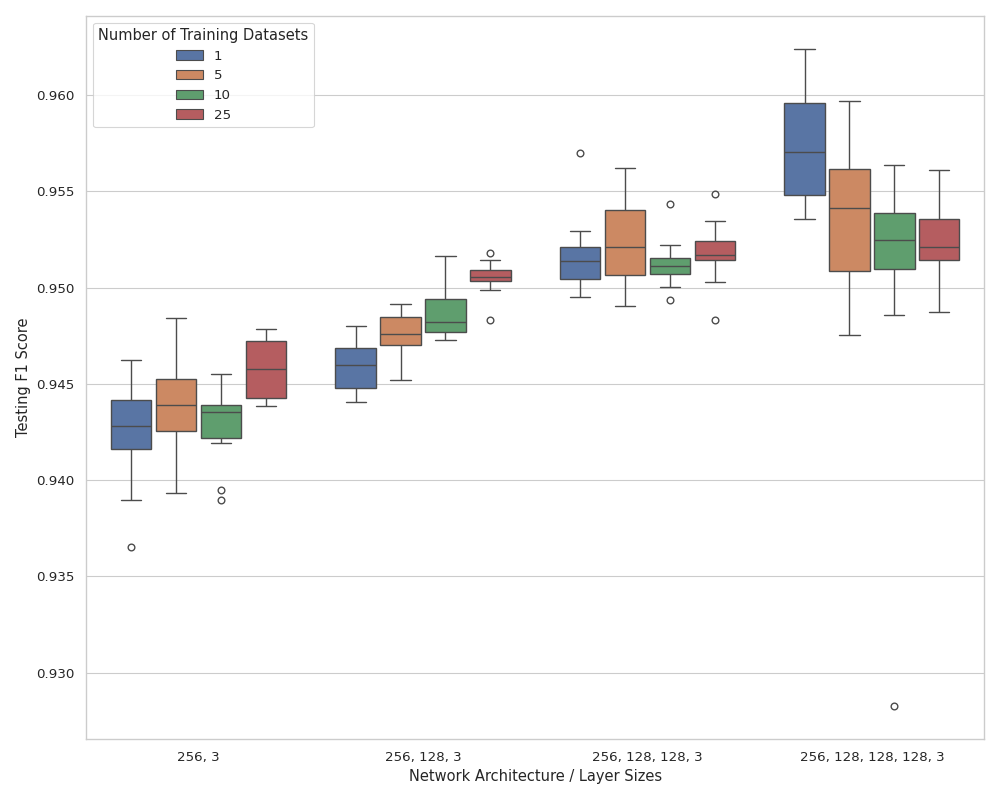}
        \subcaption{Non-Normalized Energy Profiles.}
        \label{Fig: Experiment One Results}
    \end{subfigure}
    \begin{subfigure}[t]{0.45\textwidth}
        \centering
        \includegraphics[width=1.0\textwidth]{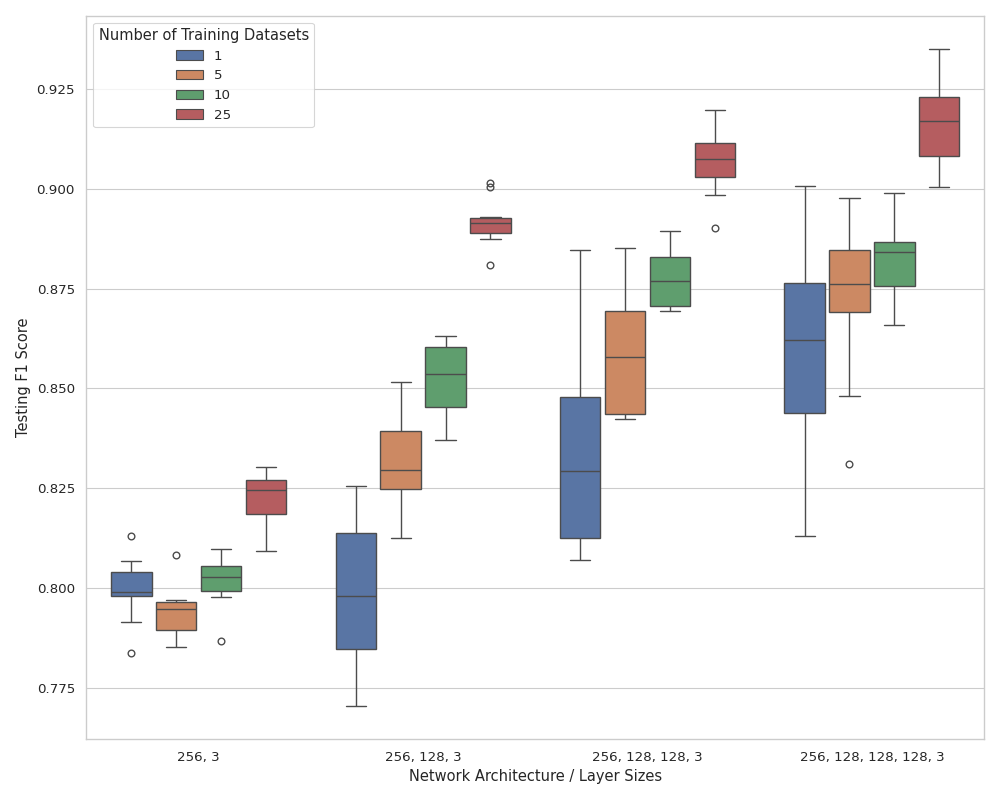}
        \subcaption{Normalized Energy Profiles.}
        \label{Fig: Experiment One Results Normalized}
    \end{subfigure}
    \caption{Macro F1 score of neural networks trained on energy profiles from standard Hopfield conditions. Note the two plots use different vertical axis limits. The number of Hopfield networks used for training is shown by the color of the box plots, and layer sizes are shown alone the x-axis. Each combination of parameters (visualized as a separate box) is repeated ten times.}
\end{figure}

We start our experiments by investigating how the depth of a neural network and the number of Hopfield networks used for training impacts the testing macro F1 score. Figure \ref{Fig: Experiment One Results} shows these results for classifiers trained on non-normalized energy profiles. A deeper neural network is able to learn more complex mappings from energy profile to class, and hence may be able to perform better than shallower classifiers, and that is exactly what we see. Although deeper classifiers show higher macro F1 scores than shallower ones, it is somewhat surprising how well the shallowest classifier performs. The deepest classifier here only improves the F1 score by around \(0.01\) compared to the shallowest. For this reason we have decided to use the shallowest architecture for the remaining experiments --- there is minimal trade-off in F1 score, and it allows us to interpret the learned weights directly. We would expect that a greater number of Hopfield networks used for training would improve testing F1 score, however this trend is not as significant as predicted. Most architectures seem to perform as well with one dataset as with twenty-five, and the deepest classifier performs significantly worse. This is likely due to the deeper networks over-fitting the training data thanks to the increased complexity of the model --- smaller classifiers appear to generalize better as they are constrained in their power.

Figure \ref{Fig: Experiment One Results Normalized} shows the same experiment as Figure \ref{Fig: Experiment One Results} but using normalized energy profiles as training data.  Clearly it is much easier to distinguish the classes of the non-normalized energy profiles than the normalized ones, as the macro F1 score is significantly lower in this experiment. The trend of increased training data volume leading to higher F1 scores is exaggerated here, indicating that normalized energy profiles are more difficult for the classifier, and more training data is required to find hypotheses that separate the classes robustly. This is corroborated by the increasing F1 scores as the classifiers grow deeper --- complex mappings are required in this dataset which are facilitated by the deeper networks. The shallowest classifier has an F1 score around \(0.1\) below that of the deepest, a much wider gap than the non-normalized energy profiles in Figure \ref{Fig: Experiment One Results}. Although normalized energy profiles should allow better generalization this seems to come at the cost of requiring more data and more complex models for the simplest tasks.

\subsection{Experiment Two: Number of Prototypes}
\label{Section: Experiment Two}

\begin{figure}[H]
    \centering
    \includegraphics[width=0.8\textwidth]{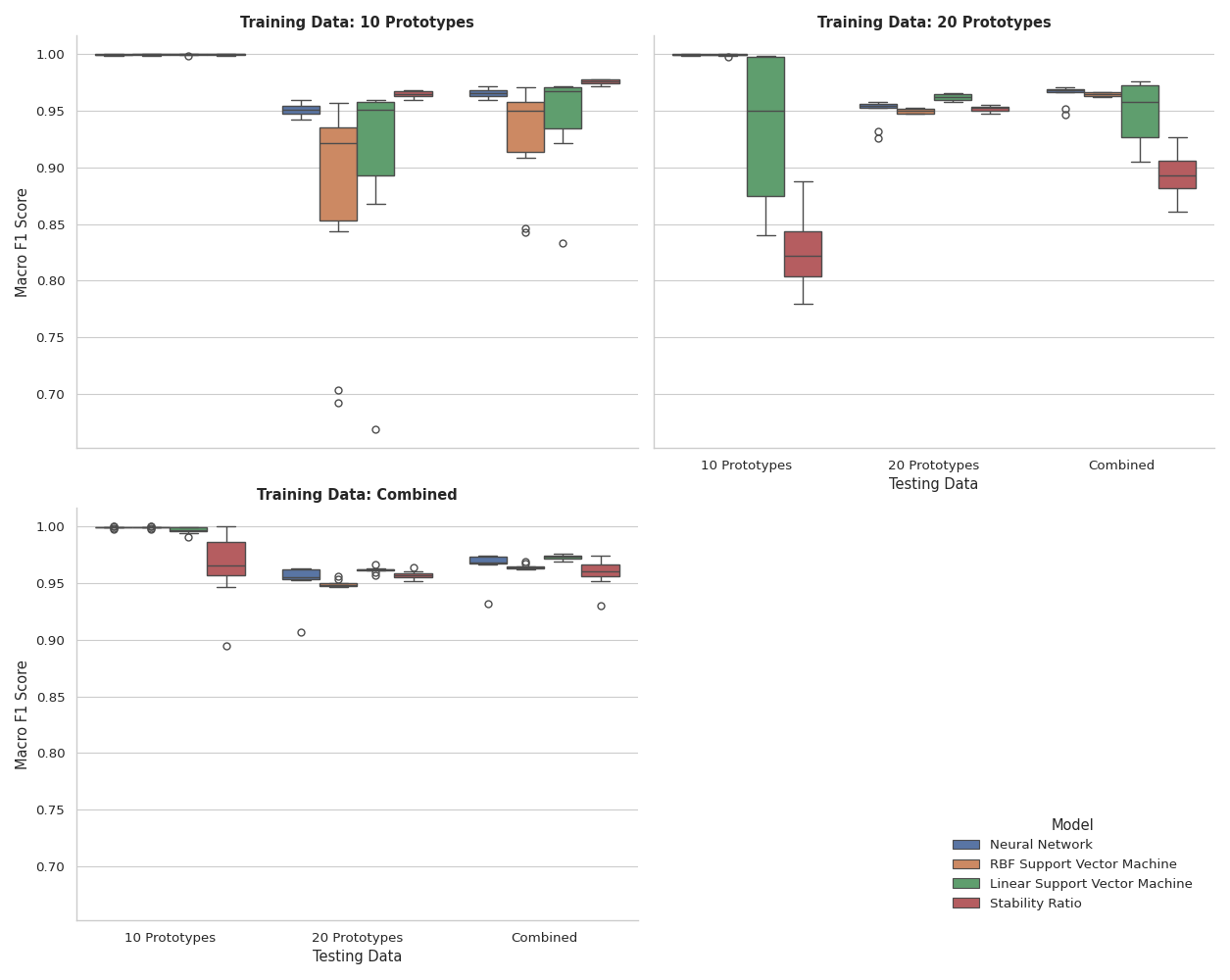}
    \caption{Testing macro F1 score of models trained on non-normalized energy profiles, varying the number of prototypes. Each combination of parameters (visualized as a separate box) is repeated ten times.}
    \label{Fig: Experiment Two Results}
\end{figure}

\begin{figure}[H]
    \centering
    \includegraphics[width=0.8\textwidth]{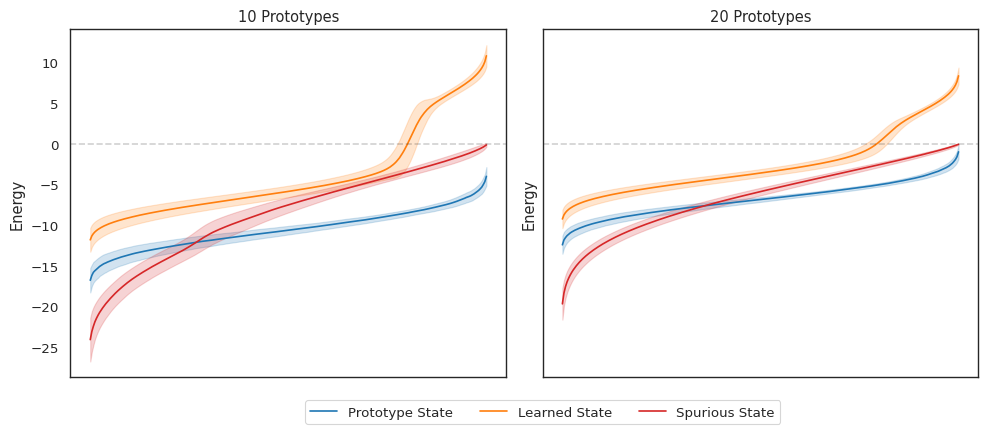}
    \caption{Sampled energy profiles of states when varying the number of prototypes learned by the Hopfield network. The mean energy profile is shown as a solid line, and the standard deviation of each neuron is shown as a shaded area around the mean.}
    \label{Fig: Experiment Two Energy Profiles}
\end{figure}

\begin{figure}[H]
    \centering
    \includegraphics[width=0.8\textwidth]{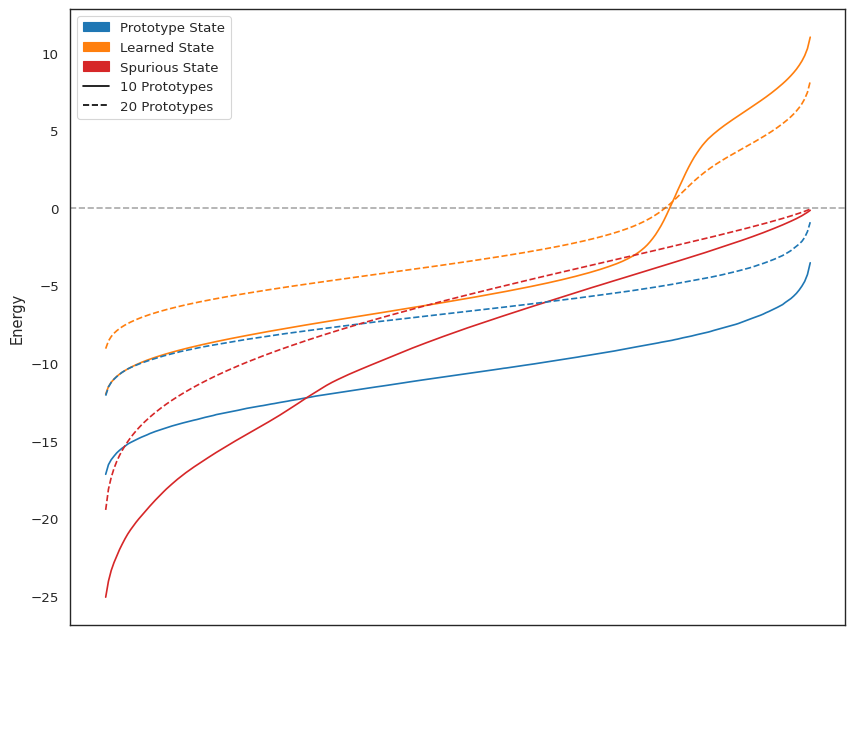}
    \caption{Energy profiles from Figure \ref{Fig: Experiment Two Energy Profiles} overlaid for easier comparison between networks with differing numbers of stored prototypes. The solid lines represent states from the ten prototype network, while dashed lines represent states from the twenty prototype network.}
\end{figure}

Altering the number of prototypes learned by the Hopfield network drastically changes the energy profiles of the learned states. Figure \ref{Fig: Experiment Two Energy Profiles} shows the energy profiles of states when 10 and 20 prototypes are learned. Note that the energy profiles of the prototype and spurious states grow more similar as the number of prototypes is increased, making these classes more difficult to distinguish. This results in more spurious states being classified as prototypes, diminishing the F1 score. Using non-normalized energy profiles the learned states are still easy to distinguish as they are unstable, so even though these energy profiles also appear to be more similar to the other classes this does not impact classifier performance. Figure \ref{Fig: Experiment Two Results} shows the testing F1 scores for classifiers trained on 10 and 20 prototypes. 

As predicted from the energy profiles in Figure \ref{Fig: Experiment Two Energy Profiles} the 10 prototype states are easy to classify --- no matter what data the classifiers were trained on, testing on 10 prototypes yields extremely high testing F1 scores except perhaps the stability ratio classifier. The 20 prototype states dataset proves slightly more difficult. Of particular interest is that all classifiers trained on 10 prototypes have a significant drop in F1 score when testing on 20 prototypes, showing that generalization is limited between Hopfield networks trained on a varying number of prototypes. 

The stability ratio classifier consistently has the lowest testing F1 score in this experiment, indicating that a slightly more complex model outperforms the existing solution. Support vector machines equipped with a linear kernel do not perform as well as we expected. For example, when training on 20 prototypes the linear SVM has a significant variance on the testing 10 prototype testing dataset, while the neural network and radial basis function SVM classifiers do not. This is likely because the energy profiles of the prototype states \textit{shift} significantly between the two classes, leading to a large number of prototype states crossing the learned decision boundary in the linear SVM which the other (non-linear) models avoid. When training on the combined dataset all of our models perform similarly, which is promising.

These results are similar to Network A and B in \citet{GormanEtAl2017}. In their work, \citeauthor{GormanEtAl2017} found that introducing a second prototype (Network B) caused all probes to become ``bad''; that is, they did not stabilize on an attractor within 1000 iterations. We did not find such a restriction even when using many more prototypes. We suspect this is due to a combination of our method of generating prototype-forming datasets, using a greater number of example states per prototype, and the use of a thermal perceptron learning rule in place of the Hebbian. \citeauthor{GormanEtAl2017} used a Gaussian window to determine the probability of a neuron being ``on'', which results in learned states that are often equal to the prototype near the peak and at the tails of the Gaussian, only differing with significant probability in the intermediate neurons. Our method for generating examples adds a Bernoulli random vector to prototype state, ensuring all neurons are equally likely to deviate from the prototype state. Although learned states that are closer to the prototype may result in stronger prototype attractors \citep{McAlisterEtAl2024a} these attractors take up less of state space, potentially explaining why \citeauthor{GormanEtAl2017} found no ``good'' probes in their multi-prototype experiments. We use double the number of learned states than \citeauthor{GormanEtAl2017} (100 to their 50), which strengthens our prototype attractors in comparison. 

Finally and most significantly, \citeauthor{GormanEtAl2017} use a thermal perceptron learning rule \citep{Frean1992} while we use the Hebbian. The thermal perceptron learning rule
\begin{equation}
    \bm{W}_{ji}\left(t+1\right) = \bm{W}_{ji}\left(t\right) + \alpha\left[\xi_i\left(t+1\right) - \xi_i\left(t\right)\right] \xi_j\left(t\right) \exp\left(\frac{-\left|\sum_j \bm{W}_{ji}\left(t\right) \xi_j\left(t\right)\right|}{T}\right)
\end{equation}
for learning rate \(\alpha\) and temperature \(T\). The thermal perceptron rule is an iterative learning rule that corrects errors in the retrieval of states by updating weights based on how wrong the retrieval was. The thermal perceptron learning rule gives a higher capacity than the Hebbian in the Hopfield network, but the iterative nature is not conducive to prototype formation, as prototype attractors are weakened by successive updates since they are seen as incorrect retrievals of learned states. This is almost certainly the root cause of the small prototype capacity observed in \citet{GormanEtAl2017}, and prototype analysis of iterative learning rules like the thermal perceptron or Widrow-Hoff rules \citep{WidrowHoff1960} are exceedingly difficult. Further research could be conducted into both the formal analysis of prototype formation with iterative learning rules and classifier performance on Hopfield networks trained with these rules.

\subsection{Experiment Three: Bernoulli Parameters}
\label{Section: Experiment Three}

\begin{figure}[H]
    \centering
    \begin{subfigure}[t]{0.65\textwidth}
        \centering
        \includegraphics[width=\textwidth]{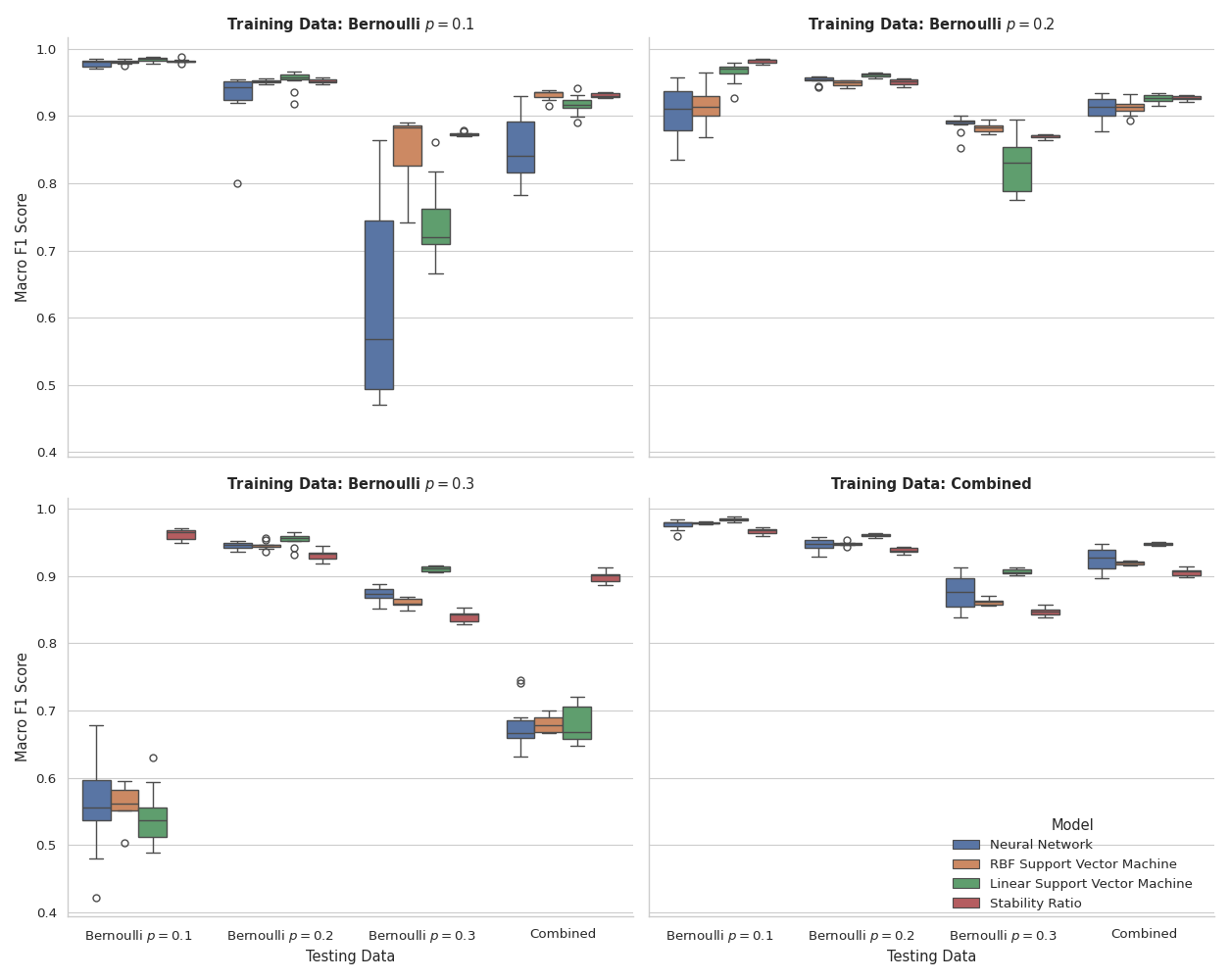}
        \subcaption{Non-Normalized Energy Profiles.}
        \label{Fig: Experiment Three Results}
    \end{subfigure}
    \begin{subfigure}[t]{0.65\textwidth}
        \centering
        \includegraphics[width=\textwidth]{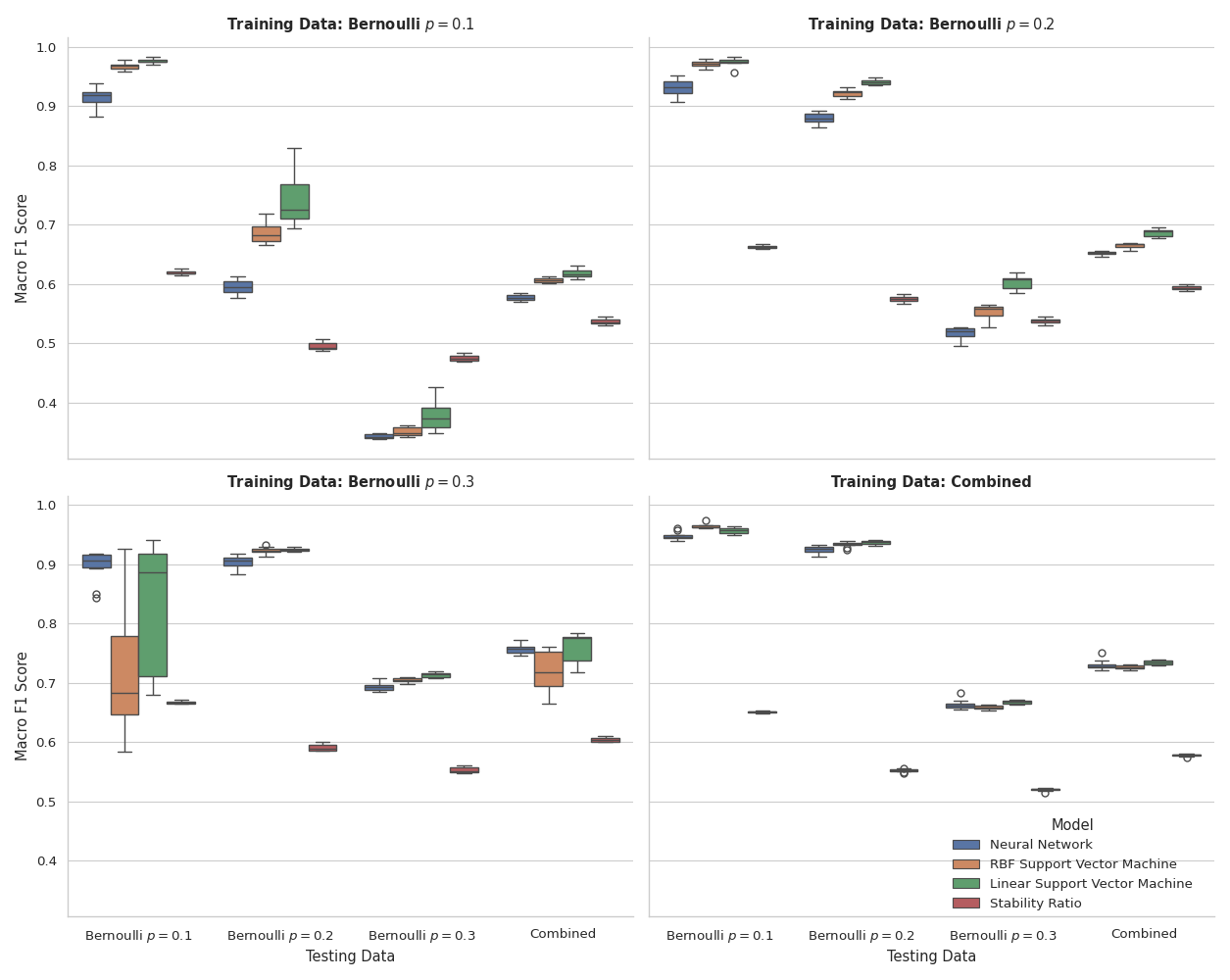}
        \subcaption{Normalized Energy Profiles.}
        \label{Fig: Experiment Three Results Normalized}
    \end{subfigure}
    \caption{Macro F1 score of models trained on energy profiles from standard Hopfield conditions, varying the Bernoulli parameter in both training and testing datasets.}
\end{figure}

\begin{figure}[H]
    \centering
    \begin{subfigure}[t]{0.45\textwidth}
        \includegraphics[width=\textwidth]{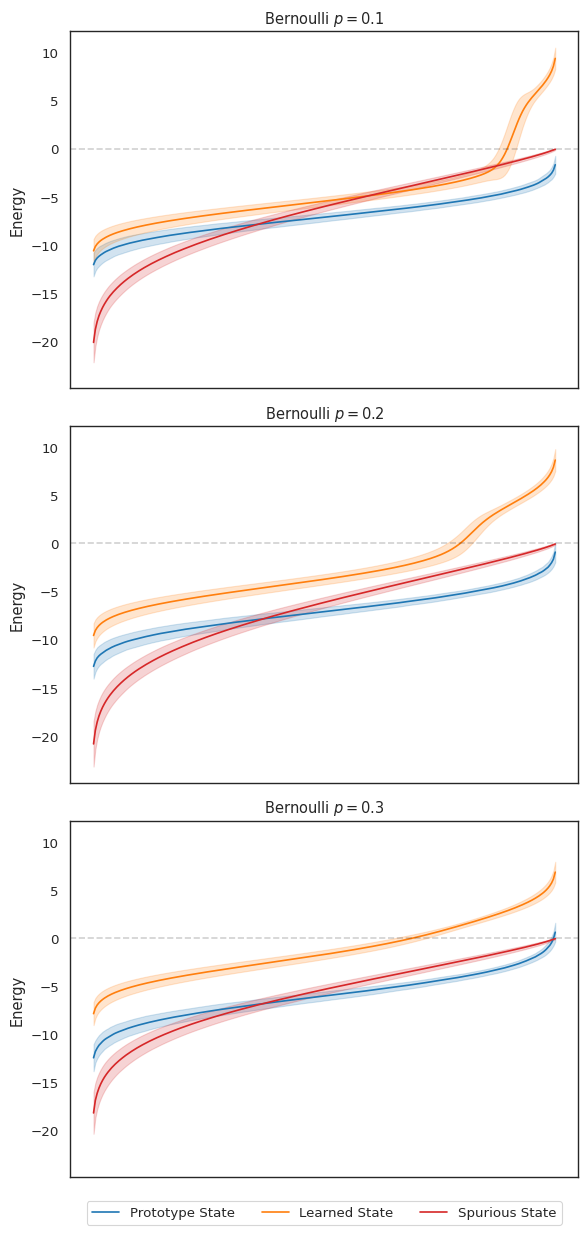}
        \subcaption{Non-normalized energy profiles.}
        \label{Figure: Experiment Three Energy Profiles Non Normalized}
    \end{subfigure}
    \begin{subfigure}[t]{0.45\textwidth}
        \includegraphics[width=\textwidth]{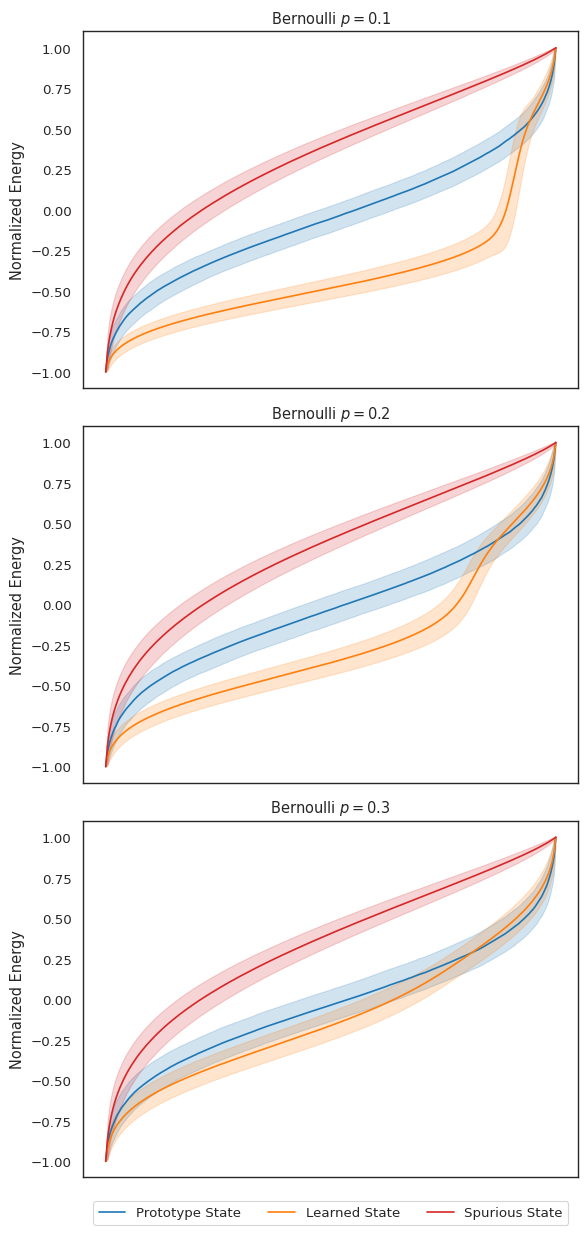}
        \subcaption{Normalized energy profiles.}
        \label{Figure: Experiment Three Energy Profiles Normalized}
    \end{subfigure}
    \caption{Energy profiles, varying the Bernoulli parameters of training data. }
\end{figure}

When varying the Bernoulli parameter of the learned states, our classifiers appear to struggle significantly. Figure \ref{Fig: Experiment Three Results} shows the testing F1 score of the classifiers trained on non-normalized energy profiles. When the Bernoulli parameter is small instances are very similar to the prototype, which induces very strong prototype formation \citep{McAlisterEtAl2024a}. Since the prototypes are formed strongly the learned states (which are necessarily nearby the prototypes) are very unstable, which separates these classes in energy profile space as seen in Figure \ref{Figure: Experiment Three Energy Profiles Non Normalized}. The spurious states are similarly distinct from the prototypes, as almost every spurious state is only barely stable in at least a few neurons, while the prototypes are strongly stable across all neurons. Therefore, a small Bernoulli parameter makes for easy classifications, exactly what we see from Figure \ref{Fig: Experiment Three Results}. When testing on a Bernoulli parameter of $p=0.1$ all classifiers do very well, except for those trained on $p=0.3$, which we discuss below. 

As the Bernoulli parameter is increased, the above arguments slowly fail --- prototypes are formed less strongly, learned states are further away from prototype attractors (and hence may be more stable), and prototype energy profiles become more similar to spurious ones. This has seemingly little impact at $p=0.2$, as all classifiers continue to learn hypotheses that successfully distinguish each class indicated by the very high macro F1 score (again, for $p\neq0.3$). Note that there is good generalization when training on $p=0.1$ and testing on $p=0.2$ data, and vice versa. Clearly these two situations are similar enough to allow for good generalization. When the Bernoulli parameter is pushed too high, in our experiments at $p=0.3$ the prototype states themselves are not stabilized. As seen in Figure \ref{Figure: Experiment Three Energy Profiles Non Normalized}, when $p=0.3$ the average prototype state is unstable. This allows our classifiers to learn a very simple hypothesis, a simple linear criterion on the most unstable neuron, and achieve fairly good results. However, although this hypothesis achieves relatively good results on the training dataset, it does not generalize very well to testing datasets that have stable prototypes, explaining the abysmal performance in this combination. The exception to this is the stability ratio classifier, which performs relatively well (only slightly below the other classifiers) but generalizes exceedingly well. This is somewhat counterintuitive considering the lessons learned from Section \ref{Section: Experiment Two}, where the stability ratio did \textit{not} handle unstable prototypes in the training data well. This could mean that the stability ratio is less sensitive to the Bernoulli parameter affecting stability than the stored prototypes approaching capacity affecting stability.

Figure \ref{Fig: Experiment Three Results Normalized} shows the same experiment varying the Bernoulli parameter but now using the normalized energy profile. We show this result in particular due to the interesting effect of noise here, specifically that it is inverted from what we observed in Figure \ref{Fig: Experiment Three Results}. When using the neural network and support vector machine classifiers with non-normalized energy profiles, increasing the noise of the training data decreased the macro F1 score on testing data; focus on the testing of the ``combined'' dataset to see this most clearly. Using normalized energy profiles, we see that increasing the noise of the training dataset \textit{increases} the macro F1 score on the testing dataset. We do not achieve F1 scores as large as with the non-normalized energy profiles, so we should refrain from using normalized energy profiles in applications, but the fact that more noise increases classifier performance is worth some thought. 

We believe this effect is due to the normalization massively increasing the similarity of each class of energy profile (see Figure \ref{Figure: Experiment Three Energy Profiles Normalized}). Our classifiers must now learn hypotheses that can separate the very similar energy profiles without exploiting knowledge of the actual stability of the states. As described above, using non-normalized energy with $p=0.3$ the energy of the most unstable neuron in prototype states became positive on average, which was an easy path for classification. Using normalized energy profiles this can no longer be exploited, so our classifiers are forced to learn the same kind of hypotheses that worked for $p=0.1,0.2$, separating on all neurons rather than just the most unstable. This would clearly generalize much better to other Bernoulli parameters, explaining the improved testing F1 score. 

When training on small Bernoulli parameters, the support vector machines outperform the neural network. We believe this is due to the maximum-margin property of the SVM classifiers. The decision boundary learned by the neural networks are only barely encompass the energy profiles of each class which leaves a lot of room misclassification when those profiles shift, either because of a different Bernoulli parameter or simply because the states are drawn from a different Hopfield network with a different distribution of prototype states. See Figure \ref{Figure: Experiment Three Energy Profiles Normalized}: the standard deviation envelope around each class for $p=0.1$ is very small compared to the distance between each class. Maximum margin models would learn to place the decision boundary halfway between those classes rather than very close to one class as learned by a gradient descent. The stability ratio classifier continues to perform consistently when using normalized energy profiles, although it is now consistently very poor.

\subsection{Number of Prototype Instances}
\label{Section: Experiment Four}

\begin{figure}[H]
    \centering
    \includegraphics[width=0.8\textwidth]{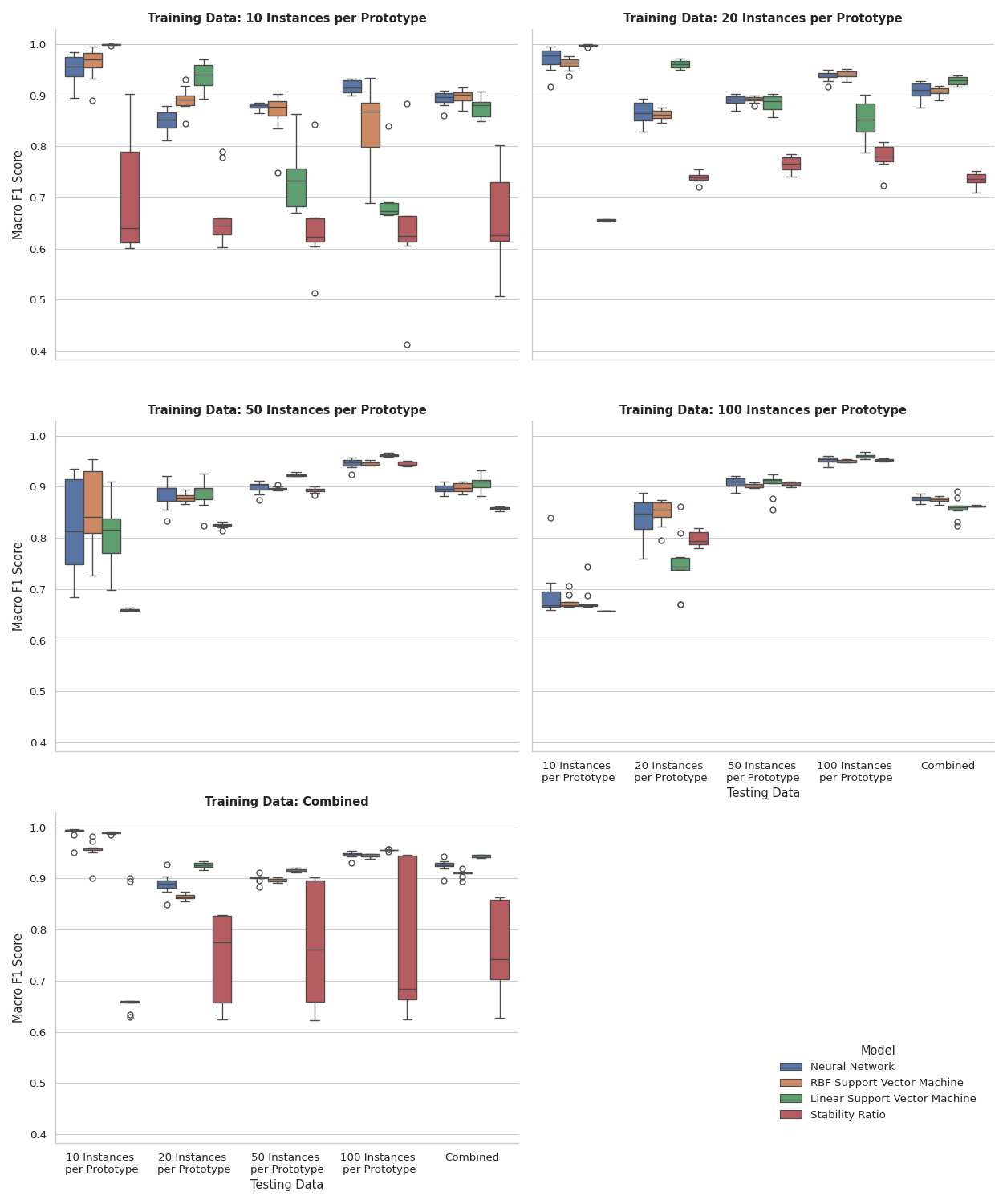}
    \caption{Macro F1 score of models trained on non-normalized energy profiles from standard Hopfield conditions, varying the number of instances per prototype.}
    \label{Fig: Experiment Four Results}
\end{figure}

\begin{figure}[H]
    \centering
    \includegraphics[width=0.8\textwidth]{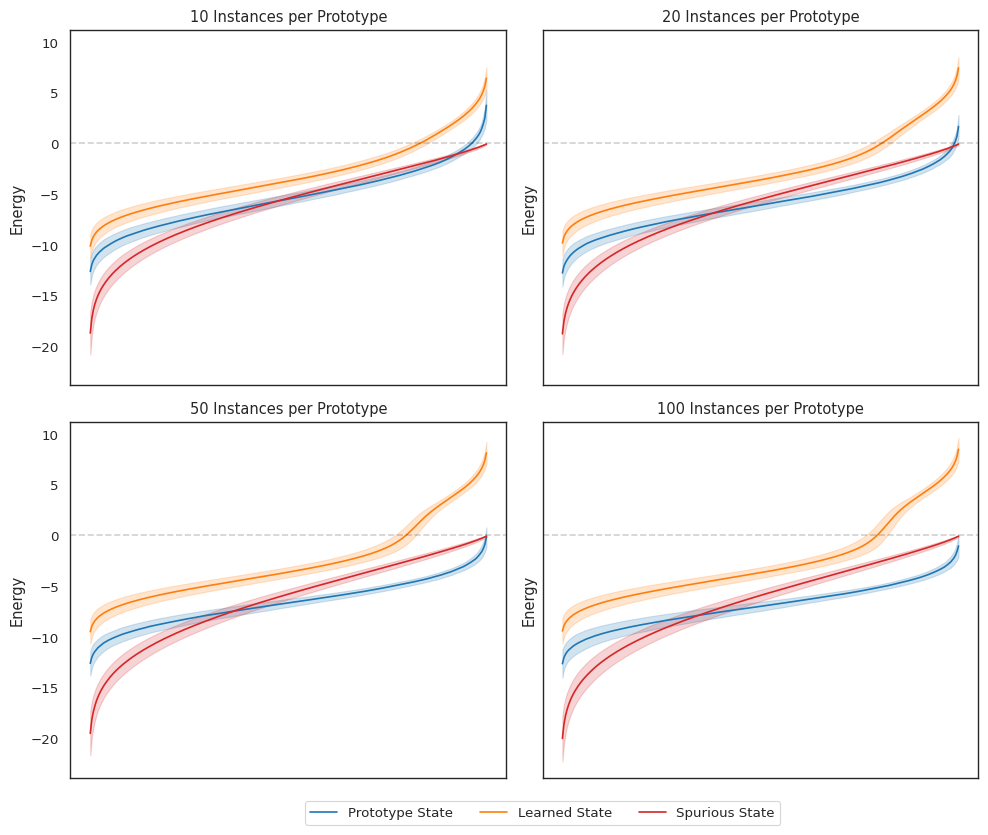}
    \caption{Sampled energy profiles of states when varying the number of instances per prototype. }
    \label{Fig: Experiment Four Energy Profiles}
\end{figure}

Figure \ref{Fig: Experiment Four Results} shows the F1 score of classifiers trained on datasets where the number of instances of each prototype is varied. When the number of instances per prototype is low, the prototype states are not stabilized strongly as the Hopfield network has not seen enough examples to form the prototype attractor in place of the individual learned states. This can be seen clearly in Figure \ref{Fig: Experiment Four Energy Profiles}, where the energy profile of prototype states is nowhere near stable at 10 instances, on average still unstable at 20 instances, and barely stable at 50 instances per prototype. Just as in Section \ref{Section: Experiment Three} the combination of unstable prototypes and non-normalized energies leads to classifiers that do not generalize well to stable prototypes. This explains why the neural network and support vector machine classifiers trained on 10 instances per prototype perform well but do not generalizer to a greater number of instances per prototype. Once a sufficient number of instances are present such that the prototype is stabilized all classifiers perform consistently well (at least, on testing data that also has strongly forming prototypes, 50 and 100 instances per prototype).  

We may question whether it is valid for a classifier to correctly distinguish an unstable prototype state from the learned states at all --- are the models with high F1 score simply over-fitting the data and identifying trends that are not useful? Is it meaningful to talk about prototypes when only 10 instances have been presented? This will likely come down to the requirements of whatever application is at hand, as even at 10 instances the prototype state is still potentially interesting for cognition; the transition from spurious to prototype is perhaps more interesting than either extreme! 

The stability ratio appears to be a very poor classifier in this experiment, particularly for unstable prototypes, but with the context of the above we could claim that it is more applicable when we want to only distinguish \textit{strongly formed} prototype states. The other models perform exceptionally well in this experiment, achieving high testing F1 scores even for the weakest prototype formation. All models have difficulty generalizing from weak to strong prototypes and vice versa, which is to be expected when the energy profiles shift as much as we see here. All models generalize well and when trained on the combined dataset, except the stability ratio classifier, which appears to have been poisoned by the weak prototypes present in the training data. In summary, the stability ratio does not handle weakly formed prototype states well, while other models are perhaps too zealous in their classification to be of use.

\subsection{Prototype-Regime and Non-Prototype-Regime Hopfield Networks}
\label{Section: Experiment Five}

\begin{figure}[H]
    \centering
    \includegraphics[width=0.8\textwidth]{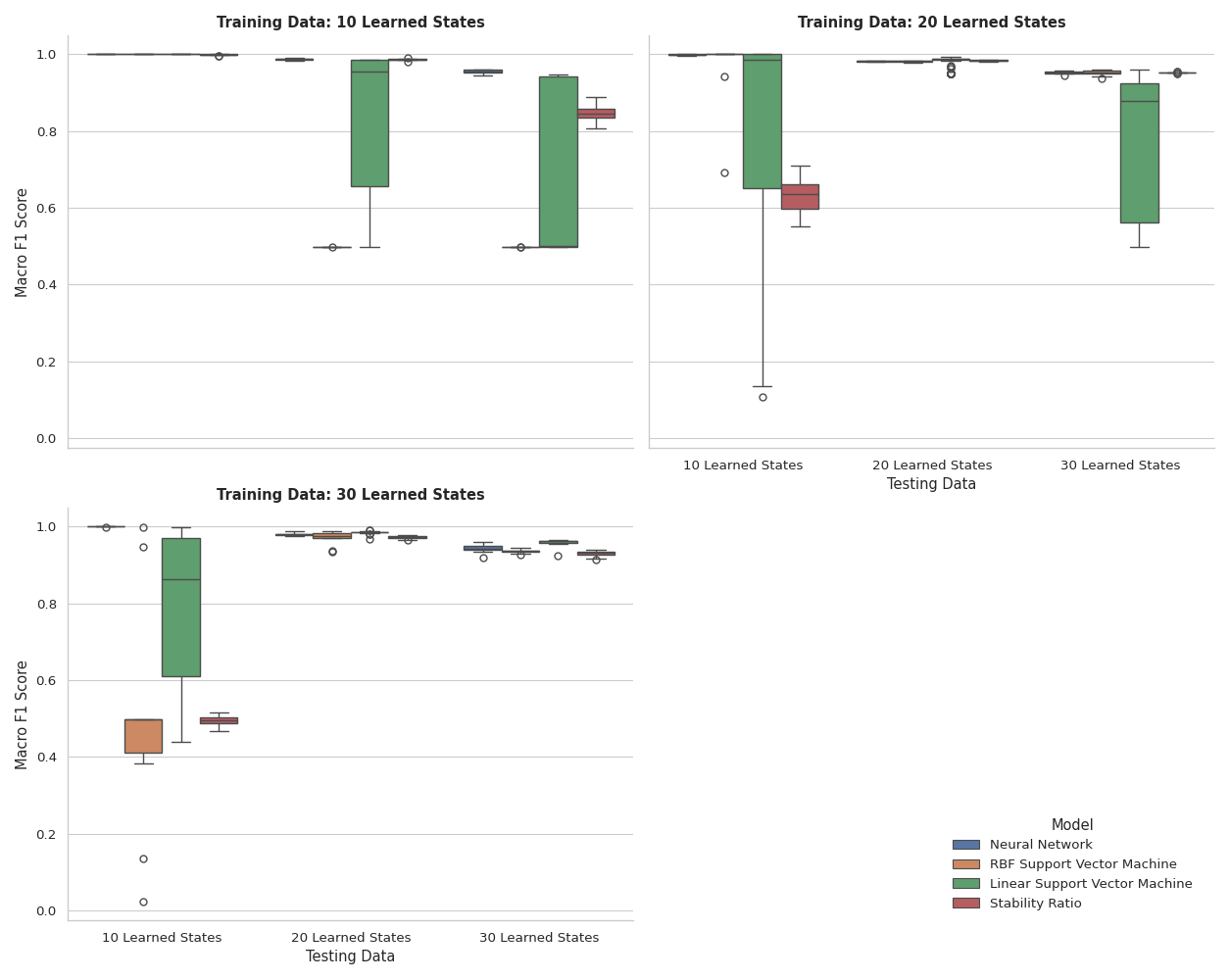}
    \caption{Macro F1 score of models trained on non-normalized energy profiles from Hopfield conditions in non-prototype-regimes.}
    \label{Fig: Experiment Five A Results}
\end{figure}

\begin{figure}[H]
    \centering
    \includegraphics[width=0.8\textwidth]{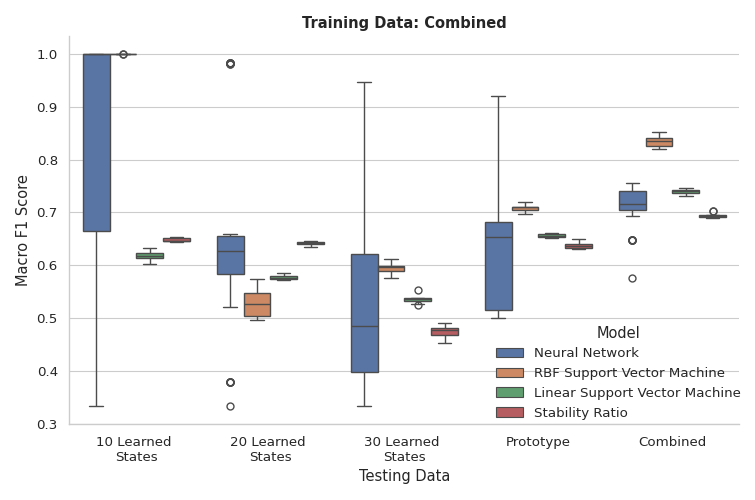}
    \caption{Macro F1 score of models trained on non-normalized energy profiles from Hopfield conditions in both the prototype- \textit{and} non-prototype-regimes.}
    \label{Fig: Experiment Five B Results}
\end{figure}

\begin{figure}[H]
    \centering
    \includegraphics[width=0.8\textwidth]{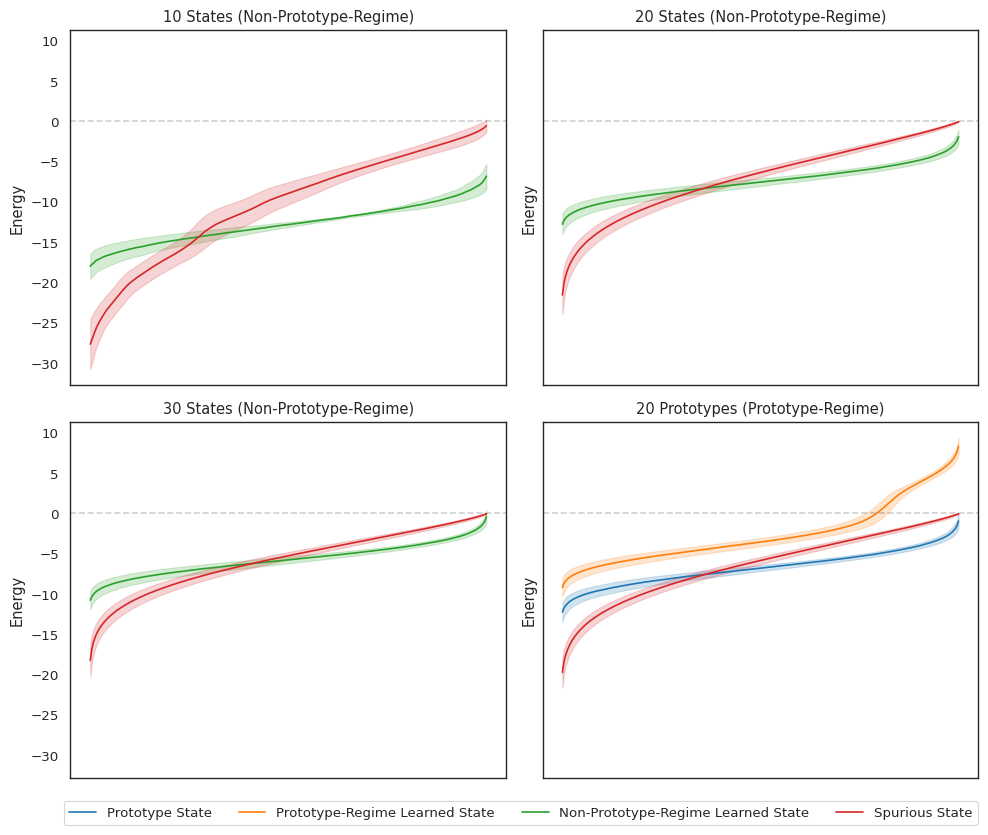}
    \caption{Sampled energy profiles of states when varying the number of learned states in non-prototype-regime Hopfield networks. }
    \label{Fig: Experiment Five Energy Profiles}
\end{figure}

In this experiment we alter our models to have 4 outputs, including the learned states for non-prototype-regime Hopfield networks as an additional class. Figure \ref{Fig: Experiment Five A Results} shows the testing F1 score for classifiers trained on non-prototype-regime Hopfield networks, i.e. Hopfield networks that learn some number of states without prototypes. This is the more typical associative task that is common to the Hopfield network. In particular, this is exactly the kind of experiment carried out by \citet{RobinsMcCallum2004} to first demonstrate the ability of the stability ratio. \citeauthor{RobinsMcCallum2004} only investigated if the stability ratio could distinguish states from a single Hopfield network at a time, rather than generalizing to multiple Hopfield networks or between different numbers of learned states. Naively this task is simpler than previous ones; we need only distinguish two classes not three. However, as seen in Figure \ref{Fig: Experiment Five Energy Profiles} those two classes have  energy profiles that are very similar as the Hopfield network nears capacity. 

We find that all classifiers perform have macro F1 scores close to 1.0 when trained and tested on a small number of learned states, far from the capacity of our \(N=256\) Hopfield network. Training on 10 learned states does not create very generalizable support vector machines, although the neural network classifier and stability ratio fair much better. Classifiers trained on both 20 and 30 learned states generalize very well to the other, although 10 learned states lags behind for the support vector machines and stability ratios. The true winner of this experiment is the neural network classifier which demonstrates very high macro F1 score across all combinations of training and testing datasets.

Perhaps more interesting, especially in the context of this paper, is the ability for our classifiers to generalize between non-prototype-regime and prototype-regime Hopfield networks. Importantly, we have decided to separate the class of prototype-regime learned states (which are almost always unstable, and are intended to be unstable) and non-prototype-regime learned states (which are intended to be stable). As seen in Figure \ref{Fig: Experiment Five Energy Profiles} the energy profiles of non-prototype-regime learned states and prototype states are extremely similar; both are flat with a small number of relatively more stable and relatively less stable neurons. As expected, the main failing of our classifiers was distinguishing these two classes. In Figure \ref{Fig: Experiment Five B Results} we present the testing F1 scores of our classifiers on each of the non-prototype-regime datasets from Figure \ref{Fig: Experiment Five A Results} as well as the ``standard'' prototype-regime Hopfield network. We only train our classifiers on the combined dataset here, as if we trained on any other combination of datasets the models would have never seen an instance of at least one class in the testing data, which would unfairly impact the F1 score.

The radial basis function support vector machine is extremely consistent on the 10 learned states dataset, managing an F1 score of 1.0 in every repetition. It appears that the energy profiles of this dataset are easy to separate, even from prototype states, mainly because the 10 learned states are so stable and have such low energy. However, in other datasets this performance drops dramatically, back to being in line with other classifiers. The neural network classifier was extremely inconsistent, often scoring both best and worst on all testing datasets. On average, it performed very poorly on the non-prototype-regime datasets, but outperformed all other models on the prototype-regime. This could be due to the class weighting used for training, perhaps weighting the classes equally is not the best approach for this task. On the combined dataset, support vector machines outperform the other classifiers, although none achieve perfect classification. The stability ratio classifier holds its own against the other classifiers and manages respectable performances considering its simplicity.
\section{Classifier Interpretability}
\label{Section: Interpretability}

We have shown that our classifiers perform and generalize as well or better than the stability ratio classifier on most prototype-yielding datasets, and now we turn our attention to interpretability.

\begin{figure}[H]
    \centering
    \begin{subfigure}[t]{0.45\textwidth}
        \includegraphics[width=\textwidth]{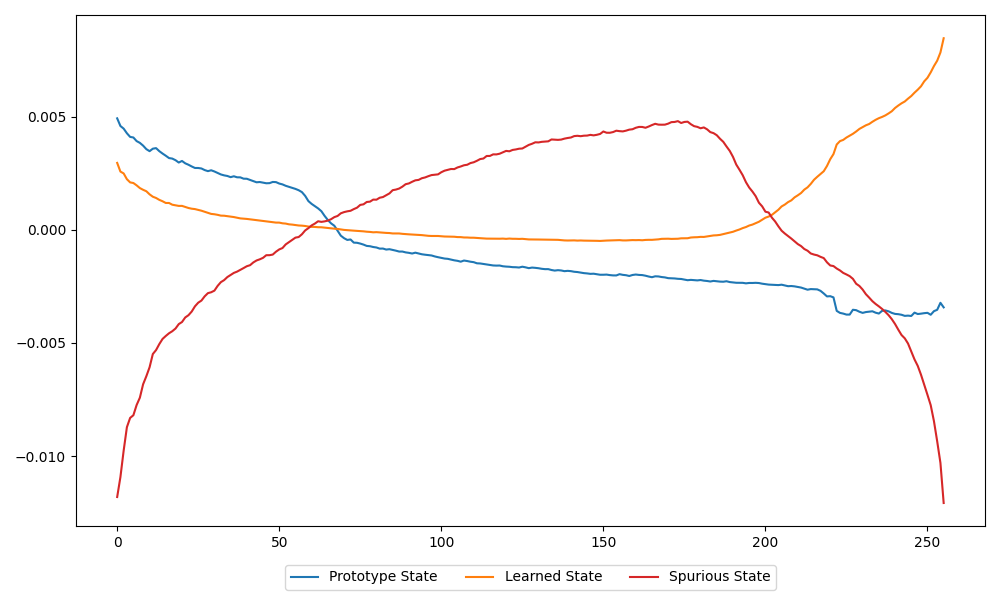}
        \subcaption{10 Prototypes.}
        \label{Figure: Linear SVM Coefficients 10 Prototype}
    \end{subfigure}
    \begin{subfigure}[t]{0.45\textwidth}
        \includegraphics[width=\textwidth]{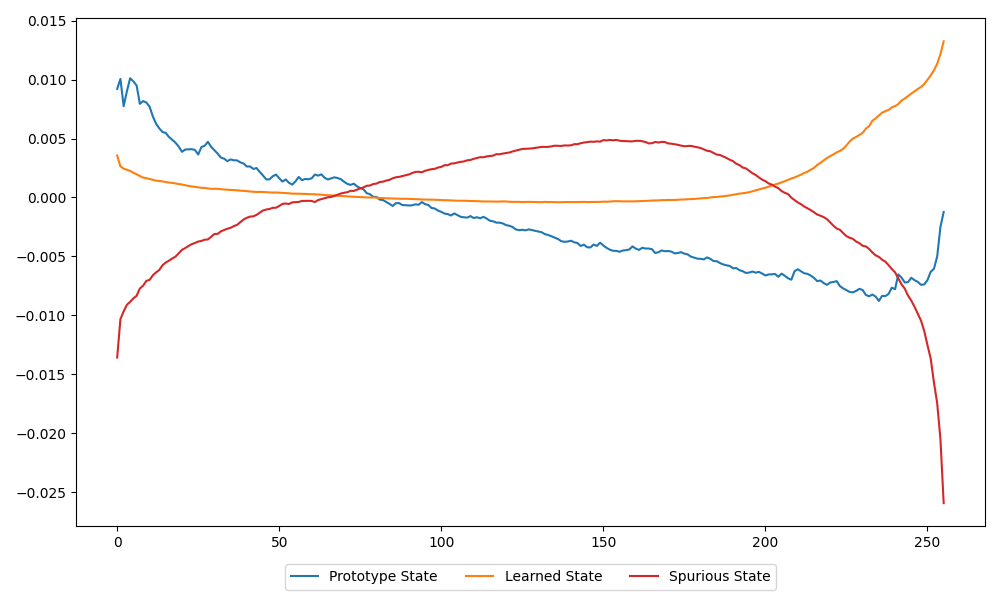}
        \subcaption{20 Prototypes.}
        \label{Figure: Linear SVM Coefficients 20 Prototype}
    \end{subfigure}
    \caption{Coefficients learned by the Linear SVM classifier trained on states from Hopfield networks learning prototype-yielding datasets. Since these classifiers were trained in a ``One-Versus-Rest'' manner the coefficients have a one-to-one mapping to each class of state. The bias term is not shown.}
    \label{Figure: Linear SVM Coefficients}
\end{figure}

\begin{figure}[H]
    \centering
    \begin{subfigure}[t]{0.45\textwidth}
        \includegraphics[width=\textwidth]{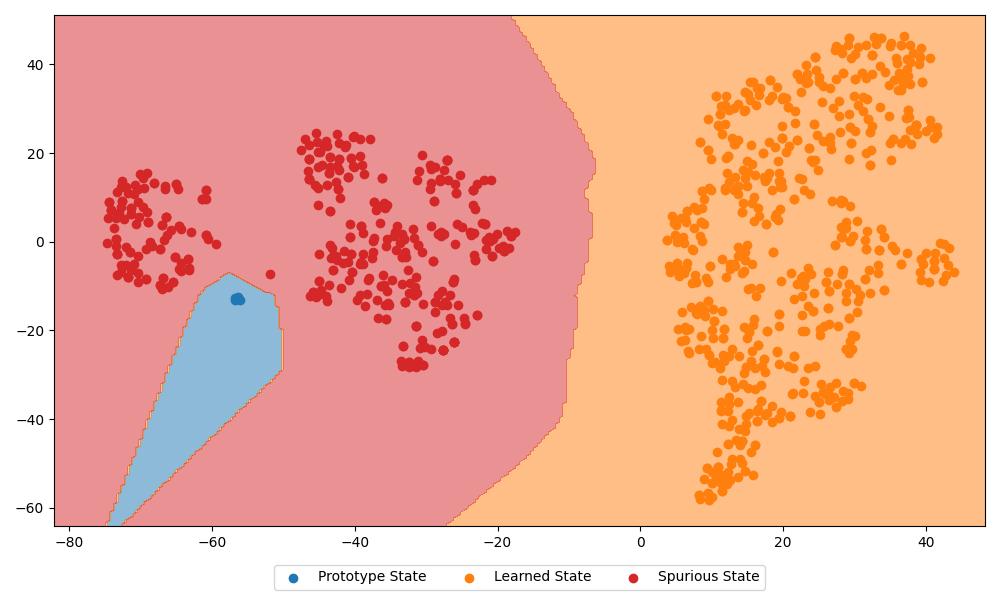}
        \subcaption{10 Prototypes.}
        \label{Figure: Linear SVM TSNE 10 Prototype}
    \end{subfigure}
    \begin{subfigure}[t]{0.45\textwidth}
        \includegraphics[width=\textwidth]{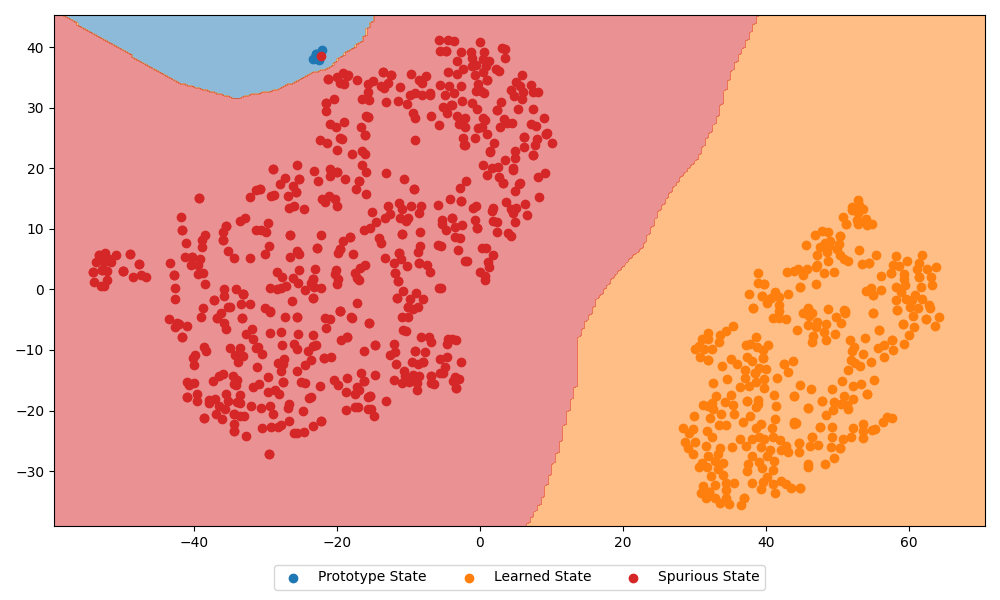}
        \subcaption{20 Prototypes.}
        \label{Figure: Linear SVM TSNE 20 Prototype}
    \end{subfigure}
    \caption{t-SNE embeddings of energy profiles, with decision boundaries of the Linear SVM shown as solid background colors.}
    \label{Figure: Linear SVM TSNE}
\end{figure}

Like the stability ratio classifier, the support vector machine equipped with a linear kernel can handle linear separable data, but unlike the stability ratio classifier the linear SVM operates over the entire energy profile rather than a single derived statistic. As we have seen, this can allow for a more powerful model, but can also provide interpretability via the learned coefficients of the SVM, an example of which is shown in Figure \ref{Figure: Linear SVM Coefficients}. Due to our configuration of the model (One-Versus-Rest), these coefficients are vectors that determine the decision boundary between one class and all other classes. Since we have selected a linear kernel, this decision boundary is a hyperplane in energy profile space, i.e. \(N=256\) space for these experiments. 

We cannot visualize these hyperplanes directly, but we can embed the energy profiles and decision boundaries into a lower dimensional space for visualization, which may reveal information about the model. T-distributed Stochastic Neighbor Embedding (t-SNE) \citep{MaatenHinton2008} is a common technique for this embedding, preserving local structure such that similar points in a high dimensional space are kept together in the lower embedding space. Figure \ref{Figure: Linear SVM TSNE} shows the t-SNE embedding of energy profiles with decision boundaries of the linear kernel SVM, in which we can see that the learned states are isolated and far from the other classes (and hence easy to classify) while the prototype states are dangerously close to spurious states, which results in confusion between the two classes. When 20 prototypes are learned, Figure \ref{Figure: Linear SVM TSNE 20 Prototype}, some spurious states cross into the decision boundary and are classified as prototype states which we observed as degraded performance of the linear SVM classifier when the Hopfield network formed weak prototypes. 

The linear kernel is extremely unforgiving when the data is not linearly separable, so when some spurious states have energy profiles very similar to prototype states this makes for an impossible problem. Worse than impossible, since the support vector machine is sensitive to outliers (the decision boundary is set as the maximum margin between the closest points of each class), even though only a few spurious states are inseparable from the prototype states this greatly impacts the learned hyperplanes and hence the generalizability of the model. Just like the stability ratio classifier, the linear SVM performs best on strong prototypes and could be used to determine if a specific Hopfield network has strongly formed or weakly formed prototype attractors. Notably, since the linear SVM operates over the energy profiles rather than the stability ratio, a combination of these two models would allow us to investigate states that are linearly separable in one domain but not the other; that is not the case for prototypes learned with the Hebbian rule, but could be possible with other prototype-yielding datasets or learning rules.

We may also directly interpret the coefficients of the linear support vector machine in Figure \ref{Figure: Linear SVM Coefficients}. Since these vectors are the maximum-margin decision boundary between one class and all others, the magnitude of the coefficients tells us how important the energy of each neuron is in determining that boundary. For example, in Figure \ref{Figure: Linear SVM Coefficients 10 Prototype} the coefficient vector for the learned states has a very low magnitude for most neurons, but a large magnitude for the most unstable neurons (those towards the right of the plot, remembering the energy profiles are sorted), meaning the key feature for distinguishing learned states from others is the very high energy of the most unstable neurons. A less obvious interpretation that can be gleaned from this plot can be found in the coefficient vector for the spurious states. The most stable and most unstable neurons have very large negative coefficients, and the middling neurons have somewhat large positive coefficients. This appears to tell us that spurious states, compared to the nearest other states, have much lower energy for their most stable neurons, higher energy for the middling neurons, and lower energy again for the least stable neurons. This may be counterintuitive at first glance --- looking at Figure \ref{Fig: Example Energy Profiles} it appears that the nearest energy profiles are the prototype states which have \textit{lower} energy at the least stable neurons --- but reasoning about high dimensional space is tricky, and the separation at the unstable neurons is likely supported by the learned states rather than the prototype states, something that we may not have guessed without this model. 

When moving to 20 prototypes stored, Figure \ref{Figure: Linear SVM Coefficients 20 Prototype}, the separating hyperplanes change slightly to reflect the change in energy profiles. The separation of spurious states and learned states has increased, and so the magnitude of the coefficients for the least stable neurons in the spurious state vector has also increased --- the linear SVM has found a greater margin between the classes. The separation of prototype states and spurious states appears to have remained the same when looking at the spurious state vector only, however cross-referencing with the prototype vector it is clear something is afoot. The much noisier prototype vector indicates the model is struggling to separate the classes, something we know is impossible since some spurious states have energy profiles nearly identical to prototype states which the linear kernel cannot handle. The invariant coefficients of the most stable neurons in the spurious state vector tells us that the vast majority of the spurious state energy profiles have not moved very much, only a relatively small number of spurious states are now nearly identical to prototype energy profiles, but the quantity is not great enough to adjust the learned hyperplane significantly. 

\begin{figure}[H]
    \centering
    \begin{subfigure}[t]{0.45\textwidth}
        \includegraphics[width=\textwidth]{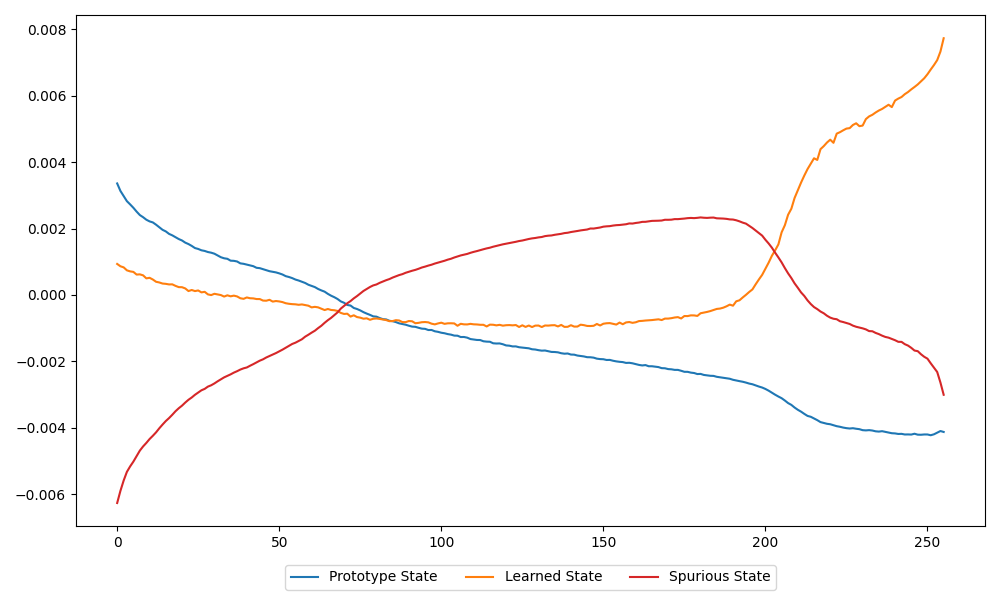}
        \subcaption{10 Prototypes.}
        \label{Figure: Neural Network Weights 10 Prototype}
    \end{subfigure}
    \begin{subfigure}[t]{0.45\textwidth}
        \includegraphics[width=\textwidth]{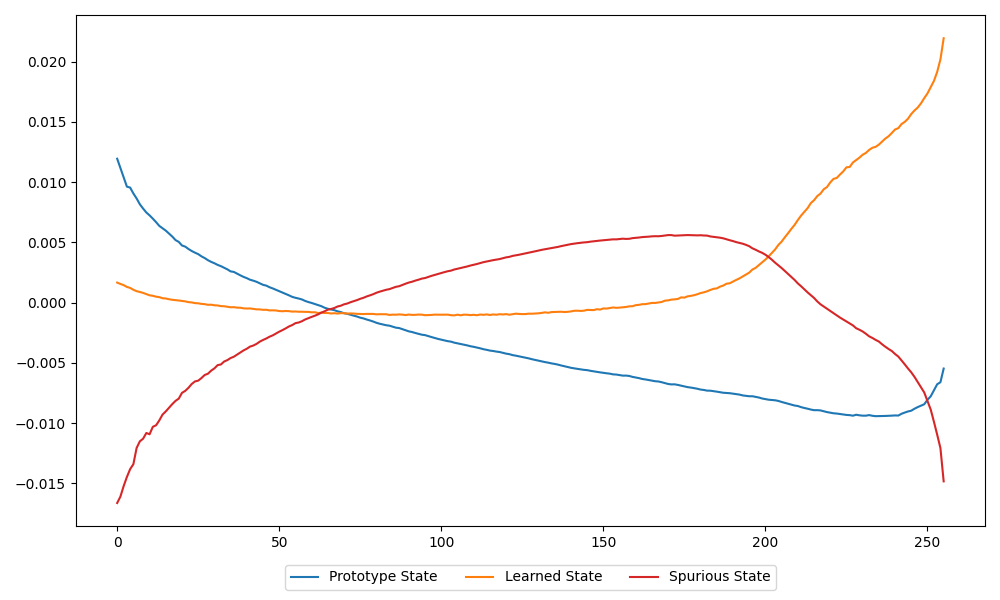}
        \subcaption{20 Prototypes.}
        \label{Figure: Neural Network Weights 20 Prototype}
    \end{subfigure}
    \caption{Parameters learned by the shallow neural network classifier trained on states from Hopfield networks learning prototype-yielding datasets. The bias term is not shown.}
    \label{Figure: Neural Network Weights}
\end{figure}

\begin{figure}[H]
    \centering
    \begin{subfigure}[t]{0.45\textwidth}
        \includegraphics[width=\textwidth]{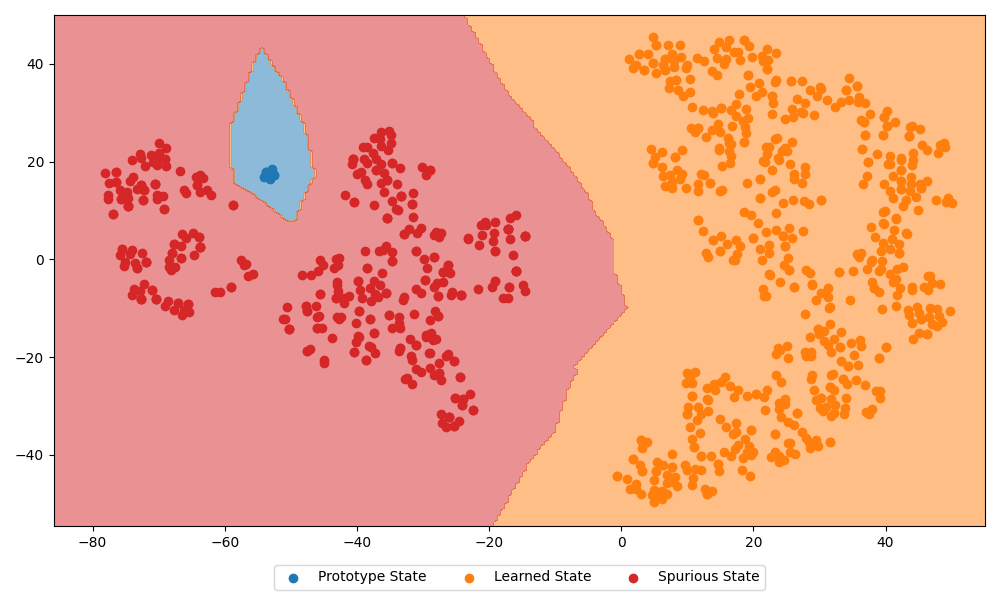}
        \subcaption{10 Prototypes.}
        \label{Figure: Neural Network TSNE 10 Prototype}
    \end{subfigure}
    \begin{subfigure}[t]{0.45\textwidth}
        \includegraphics[width=\textwidth]{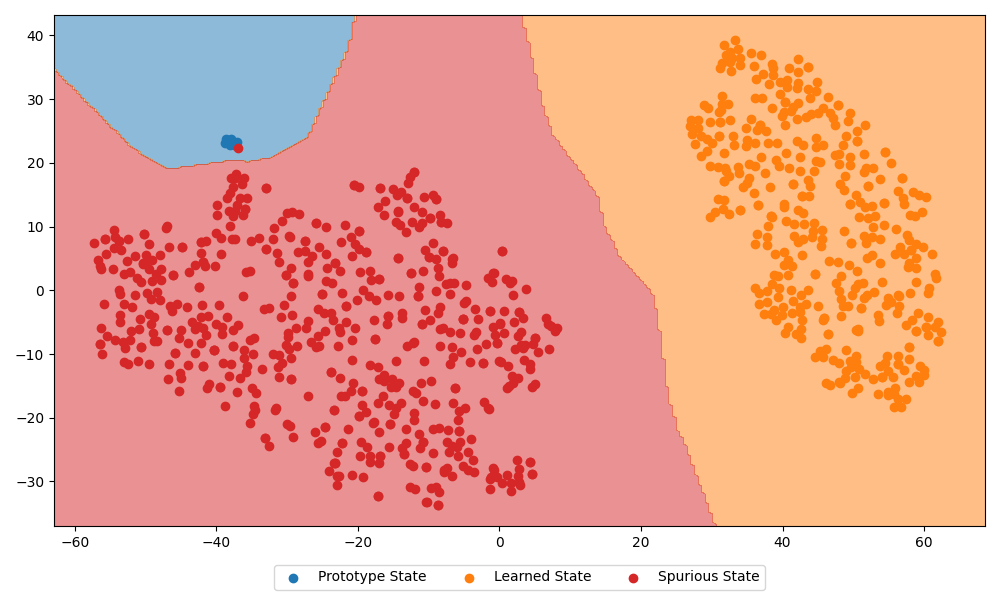}
        \subcaption{20 Prototypes.}
        \label{Figure: Neural Network TSNE 20 Prototype}
    \end{subfigure}
    \caption{t-SNE embeddings of energy profiles, with decision boundaries of the shallow neural network shown as solid background colors.}
    \label{Figure: Neural Network TSNE}
\end{figure}

Our choice of a shallow neural network also allows for direct interpretation of the learned weights. Figure \ref{Figure: Neural Network Weights 10 Prototype} shows the parameters of a shallow neural network when trained on states from a Hopfield network with 10 or 20 prototypes. Like the linear kernel support vector machine, the model has a vector of parameters for each class, but now the parameters are used to measure a similarity to the energy profiles rather than a decision boundary hyperplane. However, since the neural network is aiming to predict large \textit{positive} logits, when the presented energy profile is negative the corresponding vector is learned to be inverted. This is why the vector for learned states appears to match the energy profile of the learned states, while the vector for the prototype states is flipped relative to the prototype energy profile --- learned states have positive energy nearly everywhere in our experiments while prototype states have negative energy nearly everywhere. 

Spurious states are again slightly difficult to interpret; we must look at the vector in the context of distinguishing spurious states from the other classes, in which case the extreme negative parameters for the most stable neurons is by far the most dominant feature. The relatively large negative parameters for the least stable neurons interact minimally with spurious energy profiles due to the very small energies at these neurons, but will filter away prototype energy profiles which have a significant negative energy there. Looking at 20 prototypes network, Figure \ref{Figure: Neural Network Weights 20 Prototype}, we see a similar trend as in the linear SVM. The learned states parameter vector is mostly unchanged (these energy profiles did not change between 10 and 20 prototypes) while the prototype and spurious states parameters are more exaggerated. The prototype energy parameter vector now has a distinct ``flick'' at the least stable neurons, which may be an attempt to separate the energy profiles of prototype states and those spurious states that have become very similar to prototypes, although Figure \ref{Figure: Neural Network TSNE 20 Prototype} shows this is unsuccessful. 

\subsection{Dense Associative Memory as a Classifier}
\label{Section: DAM Classifier Interpretability}

\begin{figure}[H]
    \centering
    \begin{subfigure}[t]{0.45\textwidth}
        \includegraphics[width=\textwidth]{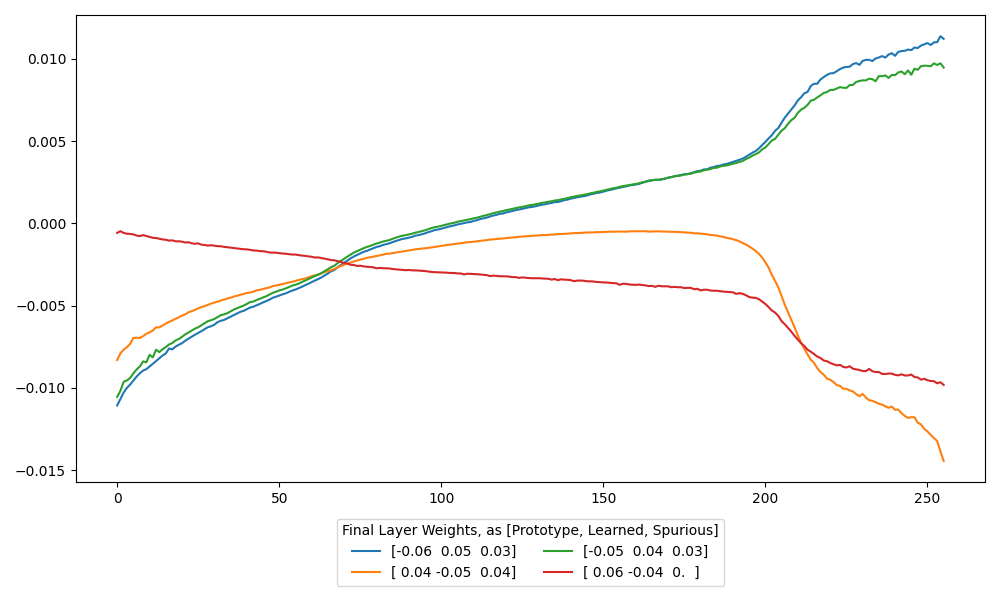}
        \subcaption{10 Prototypes.}
        \label{Figure: Deep Neural Network Weights 10 Prototype}
    \end{subfigure}
    \begin{subfigure}[t]{0.45\textwidth}
        \includegraphics[width=\textwidth]{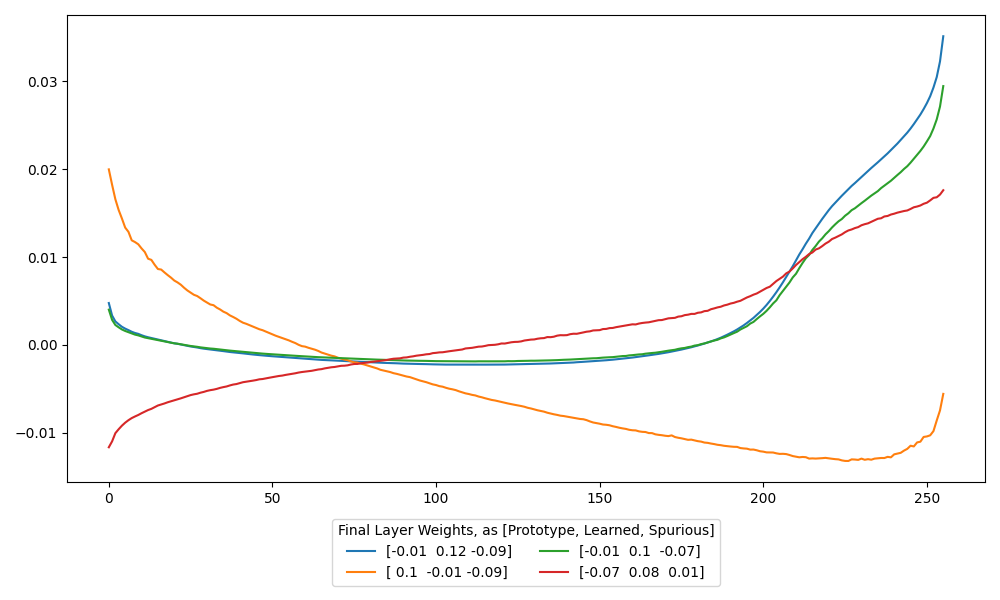}
        \subcaption{20 Prototypes.}
        \label{Figure: Deep Neural Network Weights 20 Prototype}
    \end{subfigure}
    \caption{Memory Vectors learned by the Dense Associative Memory classifier trained on states from Hopfield networks learning prototype-yielding datasets.}
    \label{Figure: Deep Neural Network Weights}
\end{figure}

The Dense Associative Memory \citep{KrotovHopfield2016} abstracts the Hopfield network to arbitrary energy functions and massively expands the capacity of the associative memory. While the exact mechanisms of the Dense Associative Memory are still nebulous \citep{McAlisterEtAl2024b}, one well known property is the feature-to-prototype transition of the memory vectors. While we have not investigated the prototype states of the Dense Associative Memory in this work, we may still use the model as a classifier for states of the Hopfield network. Specifically, as proposed by \citet{KrotovHopfield2016}, the Dense Associative Memory when used as a classifier is equivalent to a single hidden layer feed forward network, which is a slightly more powerful model than the shallow neural network we studied throughout Section \ref{Section: Results} \citep{HornikEtAl1989}. The weights of the Dense Associative Memory are still mildly interpretable: weights between the inputs and hidden layer encode features / prototypes that learn to be similar to the energy profiles, while weights between the hidden layer and outputs measure how influential the corresponding feature / prototype is on a specific class. 

We train a Dense Associative Memory with 128 memory vectors in the feed-forward equivalent form, that is a neural network with 256 inputs, 128 neurons in the hidden layer, and 3 outputs with ReLU activation functions (effectively an interaction vertex of \(n=2\)). We sort the hidden neurons by the magnitude of the weights connecting to the outputs, taking only the top few influential memory vectors and plot these in Figure \ref{Figure: Deep Neural Network Weights}. Immediately we see one disadvantage of this model in comparison to the shallow neural network and linear support vector machine: there are redundant representations for the same feature, making interpretability more difficult. Should similar features be combined when measuring similarity? How disparate should two features be before they are considered different? Another issue is that we can no longer attribute a single parameter vector to a single class, instead features influence the logits of multiple classes which makes it more difficult to determine if a learned feature is derived from energy profiles of one or all of those classes. 

Looking past these problems, the features themselves \textit{are} useful in analyzing classifier behavior. The memory vectors reflect the energy profiles of each class, as they did in the shallow neural network. Increasing the interaction vertex of the Dense Associative Memory would affect the memory vectors and could allow for more understandings of the underlying states, although this is not required in the relatively simple case we have presented in this paper. 

The Dense Associative Memory has some biologically plausible elements \citep{KrotovHopfield2021,TangKopp2021,Snow2022}, meaning this model could provide a pathway towards a biologically plausible mechanism for prototype storage, retrieval, and recognition, although much more research would be required to link these models with the correct neurological structures.
\section{Conclusion}

In this paper we investigated distinguishing the states of the Hopfield network, with a particular focus on prototype-regime Hopfield networks. Our research is a comprehensive review and extension of state classification in the Hopfield network. We extend previous work on distinguishing states of the Hopfield network \citep{RobinsMcCallum2004,GormanEtAl2017} by training a variety of classifier models on the sorted energy profile of states from the Hopfield network. We limited our focus to the most interpretable models, including shallow neural networks, linear support vector machines, and logistic regression using the stability ratio \citep{RobinsMcCallum2004}. The radial basis function support vector machine was also included to investigate the performance of a more powerful model, although the performance gain was minimal compared to the more interpretable models. We also investigated deep, densely connected, feed-forward neural networks and found that very shallow classifiers were as capable as deeper ones at distinguishing between prototype, prototype-associated, and spurious states.

A key part of this paper is the study of the generalizability of the classifiers. We aimed to find a classifier that could be trained on Hopfield networks with one dataset makeup and test well on Hopfield networks trained with another. To this end, we varied each of the parameters of prototype formation in the Hopfield network \citep{McAlisterEtAl2024a} and trained our classifiers on each variation, keeping all other parameters constant. We found the neural network classifier, linear support vector machine, and radial basis function support vector machine to generalize very well when varying the number of prototypes learned, the Bernoulli noise (for small noise parameters), and the number of instances per prototype (for a sufficient quantity of instances). We found that pushing the Hopfield network to its limit in terms of prototype learning (both large Bernoulli noise parameters and small instances per prototype) resulted in classifiers that did not generalize well. We also investigated the generalizability between prototype-regime and non-prototype-regime Hopfield networks, where we found all classifiers struggled to generalize between the two regimes. The stability ratio classifier, the simplest of our testing and the current approach to the task in literature, performed reasonably well in comparison to the more complex models, but fell behind when the prototype formation was very weak, such as for large Bernoulli noise parameters and small instances per prototype. 

We also presented the interpretable nature of the shallow neural network and linear kernel support vector machine in these tasks and described how these models could be useful in understanding the energy profiles of Hopfield network states under other conditions, both different dataset makeups and learning rules. This paper has shown that simple, interpretable models can be used alongside or in place of the stability ratio to classify states of the Hopfield network without sacrificing performance, generalizability, or interpretability.

\clearpage
\printbibliography

@article{Abe1993,
  title    = {Global Convergence and Suppression of Spurious States of the {{Hopfield}} Neural Networks},
  author   = {Abe, S.},
  year     = {1993},
  month    = apr,
  journal  = {IEEE Transactions on Circuits and Systems I: Fundamental Theory and Applications},
  volume   = {40},
  number   = {4},
  pages    = {246--257},
  issn     = {1558-1268},
  doi      = {10.1109/81.224297}
}

@misc{Agarap2019,
  title         = {Deep {{Learning}} Using {{Rectified Linear Units}} ({{ReLU}})},
  author        = {Agarap, Abien Fred},
  year          = {2019},
  month         = feb,
  number        = {arXiv:1803.08375},
  eprint        = {1803.08375},
  publisher     = {arXiv},
  doi           = {10.48550/arXiv.1803.08375},
  archiveprefix = {arXiv}
}

@article{AmitEtAl1985,
  title     = {Spin-Glass Models of Neural Networks},
  author    = {Amit, Daniel J. and Gutfreund, Hanoch and Sompolinsky, H.},
  year      = {1985},
  month     = aug,
  journal   = {Physical Review A},
  volume    = {32},
  number    = {2},
  pages     = {1007--1018},
  publisher = {American Physical Society},
  doi       = {10.1103/PhysRevA.32.1007}
}

@article{AmitEtAl1985a,
  title     = {Storing {{Infinite Numbers}} of {{Patterns}} in a {{Spin-Glass Model}} of {{Neural Networks}}},
  author    = {Amit, Daniel J. and Gutfreund, Hanoch and Sompolinsky, H.},
  year      = {1985},
  month     = sep,
  journal   = {Physical Review Letters},
  volume    = {55},
  number    = {14},
  pages     = {1530--1533},
  publisher = {American Physical Society},
  doi       = {10.1103/PhysRevLett.55.1530}
}

@inproceedings{AnselEtAl2024,
  title      = {{{PyTorch}} 2: {{Faster Machine Learning Through Dynamic Python Bytecode Transformation}} and {{Graph Compilation}}},
  shorttitle = {{{PyTorch}} 2},
  booktitle  = {Proceedings of the 29th {{ACM International Conference}} on {{Architectural Support}} for {{Programming Languages}} and {{Operating Systems}}, {{Volume}} 2},
  author     = {Ansel, Jason and Yang, Edward and He, Horace and Gimelshein, Natalia and Jain, Animesh and Voznesensky, Michael and Bao, Bin and Bell, Peter and Berard, David and Burovski, Evgeni and Chauhan, Geeta and Chourdia, Anjali and Constable, Will and Desmaison, Alban and DeVito, Zachary and Ellison, Elias and Feng, Will and Gong, Jiong and Gschwind, Michael and Hirsh, Brian and Huang, Sherlock and Kalambarkar, Kshiteej and Kirsch, Laurent and Lazos, Michael and Lezcano, Mario and Liang, Yanbo and Liang, Jason and Lu, Yinghai and Luk, C. K. and Maher, Bert and Pan, Yunjie and Puhrsch, Christian and Reso, Matthias and Saroufim, Mark and Siraichi, Marcos Yukio and Suk, Helen and Zhang, Shunting and Suo, Michael and Tillet, Phil and Zhao, Xu and Wang, Eikan and Zhou, Keren and Zou, Richard and Wang, Xiaodong and Mathews, Ajit and Wen, William and Chanan, Gregory and Wu, Peng and Chintala, Soumith},
  year       = {2024},
  month      = apr,
  pages      = {929--947},
  publisher  = {ACM},
  address    = {La Jolla CA USA},
  doi        = {10.1145/3620665.3640366},
  isbn       = {9798400703850},
  langid     = {english}
}

@article{Corbett1978,
  title    = {Universals in the Syntax of Cardinal Numerals},
  author   = {Corbett, G. G.},
  year     = {1978},
  month    = sep,
  journal  = {Lingua},
  volume   = {46},
  number   = {1},
  pages    = {61--74},
  issn     = {0024-3841},
  doi      = {10.1016/0024-3841(78)90054-2}
}

@article{CortesVapnik1995,
  title    = {Support-{{Vector Networks}}},
  author   = {Cortes, C. and Vapnik, V.},
  year     = {1995},
  journal  = {Machine Learning},
  volume   = {20},
  number   = {3},
  pages    = {273--297},
  doi      = {10.1023/A:1022627411411}
}

@article{DemircigilEtAl2017,
  title    = {On a {{Model}} of {{Associative Memory}} with {{Huge Storage Capacity}}},
  author   = {Demircigil, Mete and Heusel, Judith and L{\"o}we, Matthias and Upgang, Sven and Vermet, Franck},
  year     = {2017},
  month    = jul,
  journal  = {Journal of Statistical Physics},
  volume   = {168},
  number   = {2},
  pages    = {288--299},
  issn     = {0022-4715, 1572-9613},
  doi      = {10.1007/s10955-017-1806-y},
  langid   = {english}
}

@article{Frean1992,
  title    = {A "{{Thermal}}" {{Perceptron Learning Rule}}},
  author   = {Frean, Marcus},
  year     = {1992},
  month    = nov,
  journal  = {Neural Computation},
  volume   = {4},
  number   = {6},
  pages    = {946--957},
  issn     = {0899-7667},
  doi      = {10.1162/neco.1992.4.6.946}
}

@article{GormanEtAl2017,
  title      = {Hopfield Networks as a Model of Prototype-Based Category Learning: {{A}} Method to Distinguish Trained, Spurious, and Prototypical Attractors},
  shorttitle = {Hopfield Networks as a Model of Prototype-Based Category Learning},
  author     = {Gorman, Chris and Robins, Anthony and Knott, Alistair},
  year       = {2017},
  month      = jul,
  journal    = {Neural Networks},
  volume     = {91},
  pages      = {76--84},
  issn       = {08936080},
  doi        = {10.1016/j.neunet.2017.04.007}
}

@book{HahnEtAl1963,
  title     = {Theory and Application of {{Liapunov}}'s Direct Method},
  author    = {Hahn, Wolfgang and Hosenthien, Hans H. and Lehnigk, H.},
  year      = {1963},
  volume    = {3},
  publisher = {prentice-hall Englewood Cliffs, NJ},
}

@book{Hebb1949,
  title     = {The Organization of Behavior; a Neuropsychological Theory},
  author    = {Hebb, D. O.},
  year      = {1949},
  series    = {The Organization of Behavior; a Neuropsychological Theory},
  pages     = {xix, 335},
  publisher = {Wiley},
  address   = {Oxford, England}
}

@book{Hertz1991,
  title     = {Introduction {{To The Theory Of Neural Computation}}},
  author    = {Hertz, John A.},
  year      = {1991},
  publisher = {CRC Press},
  address   = {Boca Raton},
  doi       = {10.1201/9780429499661},
  isbn      = {978-0-429-49966-1}
}

@incollection{Homa1984,
  title     = {On the {{Nature}} of {{Categories}}},
  booktitle = {Psychology of {{Learning}} and {{Motivation}}},
  author    = {Homa, Donald},
  editor    = {Bower, Gordon H.},
  year      = {1984},
  month     = jan,
  volume    = {18},
  pages     = {49--94},
  publisher = {Academic Press},
  doi       = {10.1016/S0079-7421(08)60359-X}
}

@article{HomaEtAl1981,
  title    = {Limitations of Exemplar-Based Generalization and the Abstraction of Categorical Information.},
  author   = {Homa, Donald and Sterling, Sharon and Trepel, Lawrence},
  year     = {1981},
  month    = nov,
  journal  = {Journal of Experimental Psychology: Human Learning and Memory},
  volume   = {7},
  number   = {6},
  pages    = {418--439},
  issn     = {0096-1515},
  doi      = {10.1037/0278-7393.7.6.418},
  langid   = {english}
}

@article{Hopfield1982,
  title    = {Neural Networks and Physical Systems with Emergent Collective Computational Abilities.},
  author   = {Hopfield, J J},
  year     = {1982},
  month    = apr,
  journal  = {Proceedings of the National Academy of Sciences of the United States of America},
  volume   = {79},
  number   = {8},
  pages    = {2554--2558},
  issn     = {0027-8424},
  pmcid    = {PMC346238},
  pmid     = {6953413}
}

@article{Hopfield1984,
  title      = {Neurons with {{Graded Response Have Collective Computational Properties}} like {{Those}} of {{Two-State Neurons}}},
  author     = {Hopfield, J. J.},
  year       = {1984},
  journal    = {Proceedings of the National Academy of Sciences of the United States of America},
  volume     = {81},
  number     = {10},
  eprint     = {23632},
  eprinttype = {jstor},
  pages      = {3088--3092},
  publisher  = {National Academy of Sciences},
  issn       = {0027-8424}
}

@article{HopfieldTank1985,
  title    = {``{{Neural}}'' Computation of Decisions in Optimization Problems},
  author   = {Hopfield, J. J. and Tank, D. W.},
  year     = {1985},
  month    = jul,
  journal  = {Biological Cybernetics},
  volume   = {52},
  number   = {3},
  pages    = {141--152},
  issn     = {1432-0770},
  doi      = {10.1007/BF00339943},
  langid   = {english}
}

@article{HornikEtAl1989,
  title    = {Multilayer Feedforward Networks Are Universal Approximators},
  author   = {Hornik, Kurt and Stinchcombe, Maxwell and White, Halbert},
  year     = {1989},
  month    = jan,
  journal  = {Neural Networks},
  volume   = {2},
  number   = {5},
  pages    = {359--366},
  issn     = {0893-6080},
  doi      = {10.1016/0893-6080(89)90020-8}
}

@article{JourneEtAl2022,
  title    = {Hebbian {{Deep Learning Without Feedback}}},
  author   = {Journ{\'e}, Adrien and Rodriguez, Hector Garcia and Guo, Qinghai and Moraitis, Timoleon},
  year     = {2022},
  month    = sep,
  journal  = {arXiv.org},
  langid   = {english}
}

@article{Kempton1978,
  title      = {Category Grading and Taxonomic Relations: {{A}} Mug Is a Sort of a Cup},
  shorttitle = {Category Grading and Taxonomic Relations},
  author     = {Kempton, Willett},
  year       = {1978},
  journal    = {ResearchGate},
  langid     = {english}
}

@misc{KingmaBa2017,
  title         = {Adam: {{A Method}} for {{Stochastic Optimization}}},
  shorttitle    = {Adam},
  author        = {Kingma, Diederik P. and Ba, Jimmy},
  year          = {2017},
  month         = jan,
  number        = {arXiv:1412.6980},
  eprint        = {1412.6980},
  publisher     = {arXiv},
  doi           = {10.48550/arXiv.1412.6980},
  archiveprefix = {arXiv}
}

@article{KirkpatrickSherrington1978,
  title     = {Infinite-Ranged Models of Spin-Glasses},
  author    = {Kirkpatrick, Scott and Sherrington, David},
  year      = {1978},
  month     = jun,
  journal   = {Physical Review B},
  volume    = {17},
  number    = {11},
  pages     = {4384--4403},
  publisher = {American Physical Society},
  doi       = {10.1103/PhysRevB.17.4384}
}

@inproceedings{KrotovHopfield2016,
  title         = {Dense {{Associative Memory}} for {{Pattern Recognition}}},
  author        = {Krotov, Dmitry and Hopfield, John J.},
  year          = {2016},
  month         = sep,
  eprint        = {1606.01164},
  primaryclass  = {cond-mat, q-bio, stat},
  publisher     = {arXiv},
  doi           = {10.48550/arXiv.1606.01164},
  archiveprefix = {arXiv}
}

@article{KrotovHopfield2018,
  title    = {Dense {{Associative Memory Is Robust}} to {{Adversarial Inputs}}},
  author   = {Krotov, Dmitry and Hopfield, John},
  year     = {2018},
  month    = dec,
  journal  = {Neural Computation},
  volume   = {30},
  number   = {12},
  pages    = {3151--3167},
  issn     = {0899-7667},
  doi      = {10.1162/neco_a_01143}
}

@misc{KrotovHopfield2021,
  title         = {Large {{Associative Memory Problem}} in {{Neurobiology}} and {{Machine Learning}}},
  author        = {Krotov, Dmitry and Hopfield, John},
  year          = {2021},
  month         = apr,
  number        = {arXiv:2008.06996},
  eprint        = {2008.06996},
  primaryclass  = {cond-mat, q-bio, stat},
  publisher     = {arXiv},
  doi           = {10.48550/arXiv.2008.06996},
  archiveprefix = {arXiv}
}

@incollection{Kruschke2008,
  title     = {Models of Categorization},
  booktitle = {The {{Cambridge}} Handbook of Computational Psychology},
  author    = {Kruschke, John K.},
  year      = {2008},
  pages     = {267--301},
  publisher = {Cambridge University Press},
  address   = {New York, NY, US},
  doi       = {10.1017/CBO9780511816772.013},
  isbn      = {978-0-521-67410-2 978-0-521-85741-3}
}

@article{MaatenHinton2008,
  title    = {Visualizing {{Data}} Using T-{{SNE}}},
  author   = {van der Maaten, Laurens and Hinton, Geoffrey},
  year     = {2008},
  journal  = {Journal of Machine Learning Research},
  volume   = {9},
  number   = {86},
  pages    = {2579--2605},
  issn     = {1533-7928}
}

@article{McAlisterEtAl2024a,
  title = {Prototype {{Analysis}} in {{Hopfield Networks With Hebbian Learning}}},
  author = {McAlister, Hayden and Robins, Anthony and Szymanski, Lech},
  year = {2024},
  month = oct,
  journal = {Neural Computation},
  volume = {36},
  number = {11},
  pages = {2322--2364},
  issn = {0899-7667},
  doi = {10.1162/neco_a_01704},
  urldate = {2025-01-23}
}

@misc{McAlisterEtAl2024b,
  title         = {Sequential {{Learning}} in the {{Dense Associative Memory}}},
  author        = {McAlister, Hayden and Robins, Anthony and Szymanski, Lech},
  year          = {2024},
  month         = sep,
  number        = {arXiv:2409.15729},
  eprint        = {2409.15729},
  publisher     = {arXiv},
  doi           = {10.48550/arXiv.2409.15729},
  archiveprefix = {arXiv}
}

@article{McElieceEtAl1987,
  title    = {The Capacity of the {{Hopfield}} Associative Memory},
  author   = {McEliece, R. and Posner, E. and Rodemich, E. and Venkatesh, S.},
  year     = {1987},
  month    = jul,
  journal  = {IEEE Transactions on Information Theory},
  volume   = {33},
  number   = {4},
  pages    = {461--482},
  issn     = {1557-9654},
  doi      = {10.1109/TIT.1987.1057328}
}

@book{Neisser1967,
  title     = {Cognitive Psychology},
  author    = {Neisser, Ulric},
  year      = {1967},
  series    = {Cognitive Psychology},
  publisher = {Appleton-Century-Crofts},
  address   = {East Norwalk, CT, US}
}

@incollection{Nosofsky2011,
  title      = {The Generalized Context Model: An Exemplar Model of Classification},
  shorttitle = {The Generalized Context Model},
  booktitle  = {Formal {{Approaches}} in {{Categorization}}},
  author     = {Nosofsky, Robert M.},
  editor     = {Wills, Andy J. and Pothos, Emmanuel M.},
  year       = {2011},
  pages      = {18--39},
  publisher  = {Cambridge University Press},
  address    = {Cambridge},
  doi        = {10.1017/CBO9780511921322.002},
  isbn       = {978-0-511-92132-2}
}

@article{PersonnazEtAl1986,
  title    = {A Biologically Constrained Learning Mechanism in Networks of Formal Neurons},
  author   = {Personnaz, L. and Guyon, I. and Dreyfus, G. and Toulouse, G.},
  year     = {1986},
  month    = may,
  journal  = {Journal of Statistical Physics},
  volume   = {43},
  number   = {3},
  pages    = {411--422},
  issn     = {1572-9613},
  doi      = {10.1007/BF01020645},
  langid   = {english}
}

@article{PosnerKeele1968,
  title     = {On the Genesis of Abstract Ideas},
  author    = {Posner, Michael I. and Keele, Steven W.},
  year      = {1968},
  journal   = {Journal of Experimental Psychology},
  volume    = {77},
  number    = {3, Pt.1},
  pages     = {353--363},
  publisher = {American Psychological Association},
  address   = {US},
  issn      = {0022-1015},
  doi       = {10.1037/h0025953}
}

@misc{RamsauerEtAl2021,
  title         = {Hopfield {{Networks}} Is {{All You Need}}},
  author        = {Ramsauer, Hubert and Sch{\"a}fl, Bernhard and Lehner, Johannes and Seidl, Philipp and Widrich, Michael and Adler, Thomas and Gruber, Lukas and Holzleitner, Markus and Pavlovi{\'c}, Milena and Sandve, Geir Kjetil and Greiff, Victor and Kreil, David and Kopp, Michael and Klambauer, G{\"u}nter and Brandstetter, Johannes and Hochreiter, Sepp},
  year          = {2021},
  month         = apr,
  number        = {arXiv:2008.02217},
  eprint        = {2008.02217},
  primaryclass  = {cs, stat},
  publisher     = {arXiv},
  archiveprefix = {arXiv},
  langid        = {english}
}

@article{Randal1976,
  title      = {How {{Tall}} Is a {{Taxonomic Tree}}? {{Some Evidence}} for {{Dwarfism}}},
  shorttitle = {How {{Tall}} Is a {{Taxonomic Tree}}?},
  author     = {Randal, Robert},
  year       = {1976},
  journal    = {ResearchGate},
  langid     = {english}
}

@article{Robins1995,
  title     = {Catastrophic {{Forgetting}}, {{Rehearsal}} and {{Pseudorehearsal}}},
  author    = {Robins, Anthony},
  year      = {1995},
  month     = jun,
  journal   = {Connection Science},
  publisher = {Taylor \& Francis Group},
  doi       = {10.1080/09540099550039318},
  copyright = {Copyright Taylor \& Francis Group, LLC},
  langid    = {english}
}

@article{RobinsMcCallum1998,
  title    = {Catastrophic Forgetting and the Pseudorehearsal Solution in {{Hopfield-type}} Networks},
  author   = {Robins, A. and McCallum, S.},
  year     = {1998},
  journal  = {Connection Science},
  volume   = {10},
  number   = {2},
  pages    = {121--135},
  issn     = {0954-0091},
  doi      = {10.1080/095400998116530},
  langid   = {english}
}

@article{RobinsMcCallum2004,
  title    = {A Robust Method for Distinguishing between Learned and Spurious Attractors},
  author   = {Robins, Anthony V. and McCallum, Simon J. R.},
  year     = {2004},
  month    = apr,
  journal  = {Neural Networks: The Official Journal of the International Neural Network Society},
  volume   = {17},
  number   = {3},
  pages    = {313--326},
  issn     = {0893-6080},
  doi      = {10.1016/j.neunet.2003.11.007},
  langid   = {english},
  pmid     = {15037350}
}

@article{Rosch1973,
  title    = {Natural Categories},
  author   = {Rosch, Eleanor H.},
  year     = {1973},
  month    = may,
  journal  = {Cognitive Psychology},
  volume   = {4},
  number   = {3},
  pages    = {328--350},
  issn     = {0010-0285},
  doi      = {10.1016/0010-0285(73)90017-0}
}

@article{RoschMervis1975,
  title      = {Family Resemblances: {{Studies}} in the Internal Structure of Categories},
  shorttitle = {Family Resemblances},
  author     = {Rosch, Eleanor and Mervis, Carolyn B.},
  year       = {1975},
  journal    = {Cognitive Psychology},
  volume     = {7},
  number     = {4},
  pages      = {573--605},
  publisher  = {Elsevier Science},
  address    = {Netherlands},
  issn       = {1095-5623},
  doi        = {10.1016/0010-0285(75)90024-9}
}

@article{Ross1972,
  title      = {The Category Squish: {{Endstation Hauptwort}}},
  shorttitle = {The Category Squish},
  author     = {Ross, John Robert},
  year       = {1972},
  journal    = {ResearchGate},
  langid     = {english}
}

@book{Ross1973,
  title      = {Nouniness},
  author     = {Ross, John Robert},
  year       = {1973},
  publisher  = {MIT},
  address    = {Cambridge, Mass.},
  langid     = {english}
}

@article{Sadock2006,
  title      = {Getting {{Squishy}}},
  author     = {Sadock, Jerrold M.},
  year       = {2006},
  journal    = {Style},
  volume     = {40},
  number     = {1-2},
  eprint     = {10.5325/style.40.1-2.84},
  eprinttype = {jstor},
  pages      = {84--88},
  publisher  = {Penn State University Press},
  issn       = {0039-4238}
}

@article{scikit-learn,
  title   = {Scikit-Learn: {{Machine}} Learning in {{Python}}},
  author  = {Pedregosa, F. and Varoquaux, G. and Gramfort, A. and Michel, V. and Thirion, B. and Grisel, O. and Blondel, M. and Prettenhofer, P. and Weiss, R. and Dubourg, V. and Vanderplas, J. and Passos, A. and Cournapeau, D. and Brucher, M. and Perrot, M. and Duchesnay, E.},
  year    = {2011},
  journal = {Journal of Machine Learning Research},
  volume  = {12},
  pages   = {2825--2830}
}

@inproceedings{sklearn-api,
  title     = {{{API}} Design for Machine Learning Software: Experiences from the Scikit-Learn Project},
  booktitle = {{{ECML PKDD}} Workshop: {{Languages}} for Data Mining and Machine Learning},
  author    = {Buitinck, Lars and Louppe, Gilles and Blondel, Mathieu and Pedregosa, Fabian and Mueller, Andreas and Grisel, Olivier and Niculae, Vlad and Prettenhofer, Peter and Gramfort, Alexandre and Grobler, Jaques and Layton, Robert and VanderPlas, Jake and Joly, Arnaud and Holt, Brian and Varoquaux, Ga{\"e}l},
  year      = {2013},
  pages     = {108--122}
}

@article{SmithMinda1998,
  title      = {Prototypes in the {{Mist}}: {{The Early Epochs}} of {{Category Learning}}},
  shorttitle = {Prototypes in the {{Mist}}},
  author     = {Smith, J. David and Minda, John Paul},
  year       = {1998},
  journal    = {Journal of experimental psychology. Learning, memory, and cognition},
  volume     = {24},
  number     = {6},
  pages      = {1411--1436},
  publisher  = {American Psychological Association},
  address    = {Washington, DC},
  issn       = {0278-7393},
  doi        = {10.1037/0278-7393.24.6.1411},
  langid     = {english}
}

@article{SmithMinda1999,
  title      = {"{{Prototypes}} in the Mist: {{The}} Early Epochs of Category Learning": {{Correction}} to {{Smith}} and {{Minda}} (1998)},
  shorttitle = {"{{Prototypes}} in the Mist},
  author     = {Smith, J. David and Minda, John Paul},
  year       = {1999},
  journal    = {Journal of experimental psychology. Learning, memory, and cognition},
  volume     = {25},
  number     = {1},
  pages      = {69--69},
  publisher  = {American Psychological Association},
  issn       = {0278-7393},
  doi        = {10.1037/h0090333},
  langid     = {english}
}

@article{SmithMinda2000,
  title     = {Thirty Categorization Results in Search of a Model},
  author    = {Smith, David J. and Minda, John Paul},
  year      = {2000},
  journal   = {Journal of Experimental Psychology: Learning, Memory, and Cognition},
  volume    = {26},
  number    = {1},
  pages     = {3--27},
  publisher = {American Psychological Association},
  address   = {US},
  issn      = {1939-1285},
  doi       = {10.1037/0278-7393.26.1.3}
}

@article{Snow2022,
  title     = {Biological {{Plausibility}} in {{Modern Hopfield Networks}}},
  author    = {Snow, Mallory},
  year      = {2022},
  month     = dec,
  publisher = {University of Waterloo},
  langid    = {english}
}

@inproceedings{Storkey1997,
  title     = {Increasing the Capacity of a Hopfield Network without Sacrificing Functionality},
  booktitle = {Artificial {{Neural Networks}} --- {{ICANN}}'97},
  author    = {Storkey, Amos},
  editor    = {Gerstner, Wulfram and Germond, Alain and Hasler, Martin and Nicoud, Jean-Daniel},
  year      = {1997},
  series    = {Lecture {{Notes}} in {{Computer Science}}},
  pages     = {451--456},
  publisher = {Springer},
  address   = {Berlin, Heidelberg},
  doi       = {10.1007/BFb0020196},
  isbn      = {978-3-540-69620-9},
  langid    = {english}
}

@misc{TangKopp2021,
  title        = {A Remark on a Paper of {{Krotov}} and {{Hopfield}} [{{arXiv}}:2008.06996]},
  shorttitle   = {A Remark on a Paper of {{Krotov}} and {{Hopfield}} [{{arXiv}}},
  author       = {Tang, Fei and Kopp, Michael},
  year         = {2021},
  month        = may,
  journal      = {arXiv.org},
  howpublished = {https://arxiv.org/abs/2105.15034v2},
  langid       = {english}
}

@inproceedings{VaswaniEtAl2017,
  title     = {Attention Is All You Need},
  booktitle = {Proceedings of the 31st {{International Conference}} on {{Neural Information Processing Systems}}},
  author    = {Vaswani, Ashish and Shazeer, Noam and Parmar, Niki and Uszkoreit, Jakob and Jones, Llion and Gomez, Aidan N. and Kaiser, {\L}ukasz and Polosukhin, Illia},
  year      = {2017},
  month     = dec,
  series    = {{{NIPS}}'17},
  pages     = {6000--6010},
  publisher = {Curran Associates Inc.},
  address   = {Red Hook, NY, USA},
  isbn      = {978-1-5108-6096-4}
}

@article{WidrowHoff1960,
  title      = {Adaptive Switching Circuits},
  shorttitle = {(1960) {{Bernard Widrow}} and {{Marcian E}}. {{Hoff}}, {{Adaptive}} Switching Circuits, 1960 {{IRE WESCON Convention Record}}, {{New York}}},
  author     = {Widrow, Bernard and Hoff, Marcian E.},
  year       = {1960},
  doi        = {10.7551/mitpress/4943.003.0012},
  langid     = {english}
}

\appendix
\clearpage
\section{Stability Ratios}
\label{Appendix: Stability Ratios} 

We have visualized the stability ratios for our experiments to determine if a simple linear cutoff would generalize in our experiments. Our formalization of the energy is the negative of \citet{RobinsMcCallum2004,GormanEtAl2017}, so our stability ratios are seemingly ``upside-down'' compared to these works. Our stability ratios are also slightly different to those presented in \citet{GormanEtAl2017}, which we attribute to different prototype dataset synthesis and learning rules.

\begin{figure}[H]
    \centering
    \includegraphics[width=1.0\textwidth]{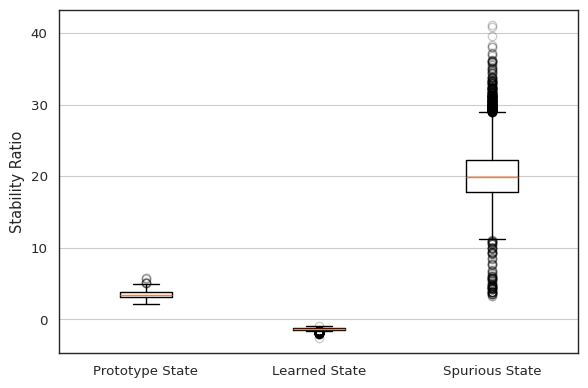}
    \caption{Stability ratio of standard condition states.}
    \label{Fig: Stability Ratio Task 01}
\end{figure}

Figure \ref{Fig: Stability Ratio Task 01} shows the stability ratio of states from the standard Hopfield conditions. Note that the stability ratio of spurious states overlaps those of prototype states for some outlier spurious states, which we observed as a significant number of spurious states being classified as prototypes, such as in Table \ref{Table: Example Confusion Matrix}; a problem that plagues all of our classifiers.

\begin{figure}[H]
    \centering
    \includegraphics[width=1.0\textwidth]{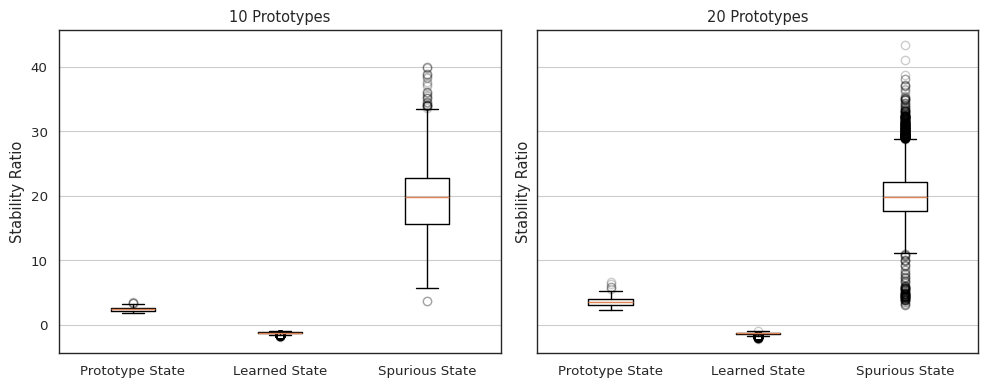}
    \caption{Stability ratio of states when varying the number of prototypes learned.}
    \label{Fig: Stability Ratio Task 02}
\end{figure}

Figure \ref{Fig: Stability Ratio Task 02} shows the stability ratios of states when the Hopfield network is trained on 10 and 20 prototypes. The maximum stability ratio of prototype states is lower and the minimum stability ratio of spurious states is higher when learning 10 prototypes than 20 prototypes, to the point that we \textit{could} draw a linear separation criterion for 10 prototypes that does \textit{not} generalize between these two datasets --- exactly the problem we aim to solve with our work.

\begin{figure}[H]
    \centering
    \includegraphics[width=0.5\textwidth]{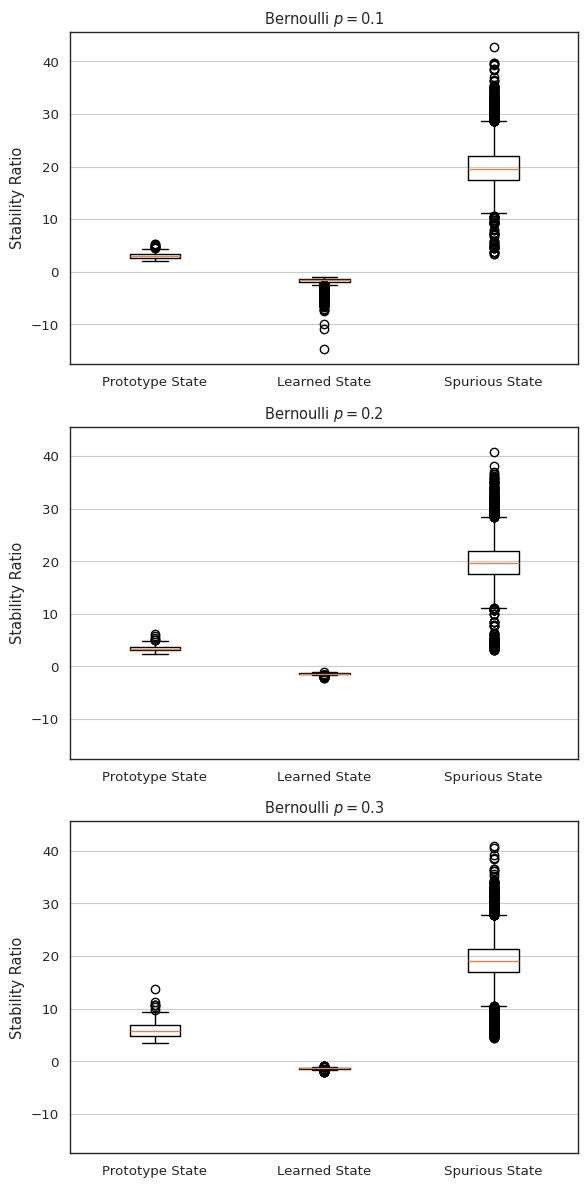}
    \caption{Stability ratio of states when varying the Bernoulli noise parameter.}
    \label{Fig: Stability Ratio Task 03}
\end{figure}

Figure \ref{Fig: Stability Ratio Task 03} shows the stability ratios as we increase the Bernoulli noise parameter of our dataset. The overlap between the stability ratios of prototype and spurious states at $p=0.1$ is much smaller than at $p=0.3$ --- as we increase the noise in our prototype dataset it becomes more difficult to distinguish between the prototype states and spurious states. A linear separation criterion developed for $p=0.1$ would likely misclassify many of the prototype states at $p=0.3$.

\begin{figure}[H]
    \centering
    \includegraphics[width=0.8\textwidth]{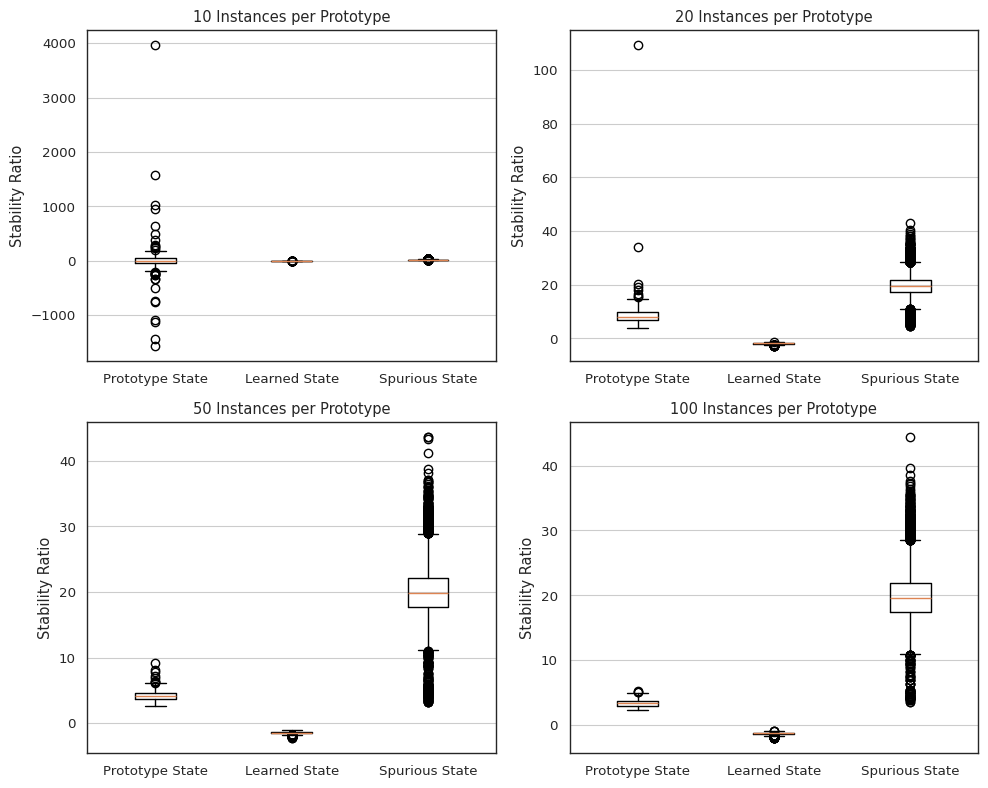}
    \caption{Stability ratio of states when varying the number of instances per prototype.}
    \label{Fig: Stability Ratio Task 04}
\end{figure}

Figure \ref{Fig: Stability Ratio Task 04} shows the stability ratios as we increase the number of instances per prototype. Note that the y-axis is not shared for this specific figure unlike our other figures, as the stability ratio of prototype states when using 10 instances per prototype is extremely varied and dominates the scale. This tells us that the prototype states do not have any special importance in the Hopfield network when trained with 10 instances per prototype; the prototypes have not formed strongly if at all. With 20 instances per prototype the prototypes have formed, but the stability ratios are still relatively varied. A linear separation criterion can not separate prototype states from spurious at 20 instances per prototype. By 50 instances per prototype the prototypes have formed very strongly and a linear criterion would perform adequately well, generalizing to higher quantities of instances per prototype too.

\begin{figure}[H]
    \centering
    \includegraphics[width=0.8\textwidth]{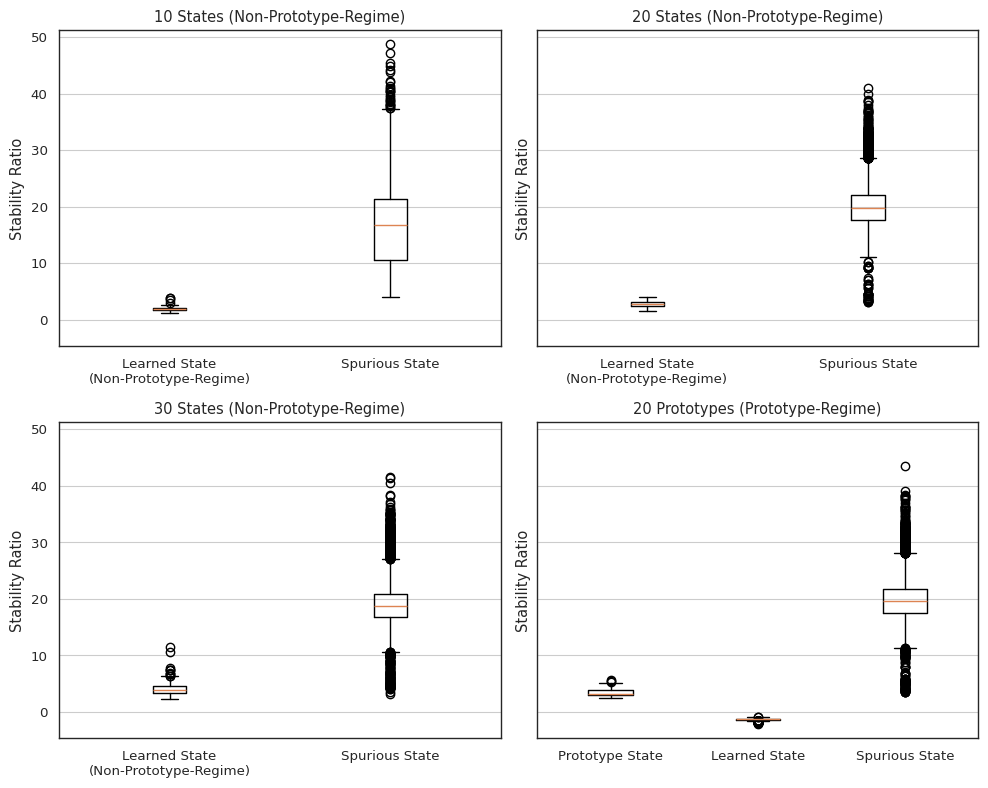}
    \caption{Stability ratio of states from both prototype and non-prototype associative memory tasks.}
    \label{Fig: Stability Ratio Task 05}
\end{figure}

Figure \ref{Fig: Stability Ratio Task 05} shows the stability ratios of states from Hopfield networks trained on both prototype tasks and non-prototype tasks (traditional associative memory tasks). Of interest here is the ability to separate prototype states in one network from regular stable attractors in another. For networks that are not near capacity (10 non-prototype-regime learned states only) this appears to be achievable with a linear separation, but even at 20 non-prototype-regime learned states only this proves impossible. Approaching the Hebbian capacity  (30 non-prototype-regime learned states) shifts the stability ratios of those states closer to the tail-end of spurious states, meaning it is increasingly difficult to distinguish learned states from spurious --- as we would expect since the learned states are approaching instability.

\end{document}